%% file: paperNN.tex
\documentclass[journal]{IEEEtran}

\usepackage{amssymb,amsmath,amsthm,amsfonts,algorithm,algorithmic,subcaption,graphicx,cite,color,flushend,mysymbol,hyperref,mathrsfs,lipsum,dsfont, multirow}
\graphicspath{{figures/}}
\newcommand \code[1] {\small\texttt{#1}\normalsize}
\def \varnet {\textsc{VarNet} }

\AtEndDocument{\par\leavevmode}

\begin{document}

\pagenumbering{gobble}

\title{VarNet: Variational Neural Networks for the\\Solution of Partial Differential Equations}

\author{Reza~Khodayi-mehr,~\IEEEmembership{Student Member,~IEEE},~and~Michael~M.~Zavlanos,~\IEEEmembership{Senior Member,~IEEE}%
\thanks{Reza Khodayi-mehr and Michael M. Zavlanos are with the Department of Mechanical Engineering and Materials Science, Duke University, Durham, NC 27708, USA, {\tt\footnotesize \{reza.khodayi.mehr, michael.zavlanos\}@duke.edu}.
This work is supported in part by the NSF award CNS-1837499.}
}

\maketitle

\begin{abstract}
In this paper we propose a new model-based unsupervised learning method, called VarNet, for the solution of partial differential equations (PDEs) using deep neural networks (NNs). Particularly, we propose a novel loss function that relies on the variational (integral) form of PDEs as apposed to their differential form which is commonly used in the literature. Our loss function is discretization-free, highly parallelizable, and more effective in capturing the solution of PDEs since it employs lower-order derivatives and trains over measure non-zero regions of space-time. Given this loss function, we also propose an approach to optimally select the space-time samples, used to train the NN, that is based on the feedback provided from the PDE residual. The models obtained using VarNet are smooth and do not require interpolation. They are also easily differentiable and can directly be used for control and optimization of PDEs. Finally, VarNet can straight-forwardly incorporate parametric PDE models making it a natural tool for model order reduction (MOR) of PDEs. 
We demonstrate the performance of our method through extensive numerical experiments for the advection-diffusion PDE as an important case-study.
\end{abstract}


\IEEEpeerreviewmaketitle

\section{Introduction} \label{sec:intro}
\input{Intro}

\section{Problem Definition} \label{sec:NN}
\input{NN}

\section{Neural Networks for Solution of PDEs} \label{sec:OT}
\input{OT}

\section{Numerical Experiments} \label{sec:sim}
\input{sim}
\section{Conclusion} \label{sec:concl}
\input{concl}



\appendices

\section{Variational Loss Function} \label{app:VarNet}
\input{VarNet}

\ifCLASSOPTIONcaptionsoff
  \newpage
\fi

\bibliographystyle{ieeetr}\bibliography{MyBibliography}

\end{document}

%% file: Intro.tex
\IEEEPARstart{D}{ynamical} systems are typically modeled using differential equations that capture their input-output behavior. Particularly, for spatiotemporal systems the underlying dynamics are modeled using partial differential equations (PDEs) whose states live in infinite-dimensional spaces. Except for some special cases, only approximate solutions to such systems can be obtained using discretization-based numerical methods. These methods typically belong to one of three main categories, namely, finite difference (FD) methods, finite element (FE) methods, and finite volume (FV) methods \cite{NMPDE2014A}. The FD methods are often restricted to simple domains whereas the FE and FV methods can be applied to more general problems.
These methods return solutions defined over a set of grid points that sample the spatiotemporal domain; the solution at any other points can be obtained by interpolating the nearby nodal values. This local treatment of the solution makes the models obtained from these methods extremely expressive but also very expensive to evaluate and store.

In this paper we follow a different approach that relies on deep neural networks (NNs) to capture the solutions of PDEs using trainable parameters that are considerably fewer than the number of grid points used in discretization-based methods.
The proposed framework, called \varnet for variational neural networks, employs a novel loss function that relies on the variational (integral) form of PDEs and is discretization-free and highly parallelizable.
Since, compared to the differential form of the PDE, the variational form contains lower order derivatives of the solution and considers segments of space-time as opposed to single points, the proposed loss function allows to learn solutions of PDEs using significantly fewer training points. Moreover, this loss function allows the NN to learn solutions of PDEs in a model-based and unsupervised way, where training of the NN parameters is guided by the PDE itself without a need for prior labeling.
We also propose a way to optimally select the points needed to train the NN, based on the feedback provided from the residual of the PDE. Using NNs to approximate solutions of PDEs makes it trivial to solve these PDEs parametrically; these parameters need only be included as additional inputs to the NN. Consequently, a great strength of \varnet algorithm is that it can also be used as a powerful model order reduction (MOR) tool for PDEs, since the resulting models are very fast to evaluate. 

\subsection{Relevant Literature} \label{sec:NNintro}
%
Using NNs to approximate solutions of PDEs can be beneficial for the following reasons: (i) their evaluation is extremely fast and thus, unlike currently available MOR methods, there is no need to compromise accuracy for speed, (ii) parallelization of the training is trivial, and (iii) the resulting model is smooth and differentiable and thus, it can be readily used in PDE-constrained optimization problems, e.g., for source identification \cite{meJ1} or control of PDEs \cite{meJ4}.
The authors in \cite[Ch. 04]{INNMDE2015YYK} provide a review of different approaches for solving PDEs using NNs.
One group of approaches utilize NNs to memorize the solution of PDEs. Particularly, they solve the PDE using a numerical method to obtain labeled training data and often utilize convolutional NNs (CNNs), being powerful image processing tools, to capture the numerical solution in a supervised learning way \cite{SPPDEPANN2017KLY}.
For instance, \cite{CNNSFA2016GLI} proposes a fluid dynamics solver that utilizes a CNN, consisting of encoding and decoding layers, to predict the velocity field in steady-state problems. It reports a stellar speedup of four orders-of-magnitude in computation time.
The authors in \cite{PDENET2017LYMD} propose PDE-NET that is capable of identifying partial differential operators as well as approximating the solution of PDEs. Note that these approaches do not replace numerical methods but rather rely on them and introduce an extra layer of approximation.

Another class of works directly discretize the domain and construct a model based on this discretization.
For instance, \cite{NSDEMRBFN2001MT} utilizes radial basis functions to construct a linear system of equations that is solved to obtain an approximate solution of the desired PDE.
There exist also another group of methods called FE-NNs which represent the governing FE equations at the element level using artifical neurons \cite{NNRFEM1994TK,FENNDE2012XWJY,FENNSDE2005RUU}. Note that both of these approaches scale with the number of discretization points and are similar in spirit to numerical methods like the FE method.

Most relevant to the approach proposed in this paper is a set of works that also directly train a NN to approximate the solution of PDEs in an unsupervised learning way. One of the early works of this kind is \cite{ANNSOPDE1998LLF} that uses the residual of the PDE to define the required loss function. In order to remove the constraints from the training problem, the authors only consider simple domains for which the boundary conditions (BCs) can be manually enforced by a change of variables. The work in \cite{NNFDSPDE2017A} elaborates more on this technique. Although these approaches attain comparable accuracy to numerical methods, they are impractical since in general enforcing the BCs is as difficult as solving the original PDE.
Following a different approach, the work in \cite{SPDEANN2013R} utilizes a constrained back-propagation algorithm to enforce the initial and boundary conditions during training. 
In order to avoid solving a constrained training problem, the authors in \cite{MLPNNNUTM2009SYM} add the constraints corresponding to BCs to the objective as penalty terms. 
Similarly, in \cite{DLASPDE2017SS} the authors focus on the solution of PDEs with high dimensions using a long short-term memory architecture and prove a convergence result for the solution as the number of trainable parameters of the NN is increased. Although a large number of samples is required to train the model, this training data is sequentially provided to the NN.
A similar approach is proposed in \cite{PINN2019RPK} that utilizes the physical laws modeled by PDEs as regularizing terms to guide the learning process, while \cite{PDLHSM2019ZZKP} uses the PDEs to define energy fields that are minimized to train CNNs to predict the PDE solutions at discrete sets of points.
Alternatively, reinforcement learning can be used to train NNs that approximate the solutions of PDEs. For instance, in \cite{GSNDE2018WJL} the actor-critic algorithm is used where the PDE residual acts as the critic.

The literature discussed above uses the PDE residual as the loss function, which contains high order derivatives and trains over measure-zero points in space-time. Consequently, adequate learning of the PDE solution requires an extremely large number of training points. Moreover, these training points are selected arbitrarily and not based on their information content. Optimal selection of training points has been extensively studied in the context of active learning to alleviate the cost of labeling data in supervised learning.
%
%
The work in \cite{ALLS2009S} presents an extensive survey of this literature including various strategies to determine the most informative queries to be made to an oracle. Often samples with the highest ambiguity are preferred; this is closely related to informative planning methods in the robotics literature \cite{meJ2, meJ3} and optimal experiment design \cite{OMMDPSI2004D}.
Here we propose an active learning method for unsupervised learning that relies on the feedback obtained from the PDE residual to optimally select the training points themselves instead of inquiring labels for them. This is closely related to mesh adaptivity methods in the FE literature where the underlying mesh is locally refined in regions with large error values \cite{IA2005HCB}.

\subsection{Contributions}
%
The contributions of this paper can be summarized as follows:
(i) Compared to methods that use NNs to approximate the solutions of PDEs, we propose a novel loss function that relies on the variational form of the PDE. The advantages of this loss function, compared to existing approaches that use the PDE residual, are two-fold. First, it contains lower order derivatives so that the solution of the PDE can be estimated more accurately.  Note that it becomes progressively more difficult to estimate a function from its higher order derivatives since differential operators are agnostic to translation. Second, it utilizes the variational (integral) form of the PDE that considers segments of space-time as opposed to single points and imposes fewer smoothness requirements on the solution. Note that variational formulations have been successfully used in the FE method for a long time \cite{FEM2012H}. 
(ii) We propose a new way to optimally select the points needed to train the NNs that is informed by the feedback obtained from the PDE residual. Our optimal sampling method considerably improves the efficiency of the training and the accuracy of the resulting models.
%
(iii) We develop the \varnet library \cite{VarNet} that uses the proposed deep learning framework to solve PDEs.
We present thorough numerical experiments using this library to demonstrate our proposed approach. Particularly, we focus on the advection-diffusion (AD) equation although the presented approach applies to any PDE that can be solved using the FE method. Note that discretization-based solutions to the AD-PDE often have stability issues for highly advective problems \cite{CPGSFAD2005HW} which are magnified when the model is reduced. This is because this reduction typically amounts to loosing high frequency content of the original model. Through simulations we demonstrate that unlike traditional MOR methods, our approach does not suffer from such instability issues.

The proposed loss function was first introduced and used to solve a robotic PDE-constrained optimal control problem in our short paper \cite{meJ4}. Compared to \cite{meJ4}, here we introduce the \varnet library that employs the same loss function but is additionally equipped with the proposed adaptive method to optimally select the NN training points and can be used not only to solve PDEs but also for MOR. Finally, compared to \cite{meJ4}, this paper presents extensive numerical experiments to validate the \varnet algorithm.

The remainder of this paper is organized as follows. In Section \ref{sec:NN} we discuss the AD-PDE and formulate the desired training problem. Section \ref{sec:OT} is devoted to the solution of PDEs using NNs where we discuss the loss function, the optimal selection of the training points, and steps to solve an AD-PDE using the \varnet library. We present numerical experiments in Section \ref{sec:sim} and finally, Section \ref{sec:concl} concludes the paper.

%% file: NN.tex

\subsection{Advection-Diffusion PDE}
%
Let $\Omega \subset \reals^d$ denote a domain of interest where $d$ is its dimension and let $\bbx \in \Omega$ denote a location in this domain and $t \in [0, T]$ denote the time variable.
Furthermore, consider a velocity vector field $\bbu : [0,T] \times \Omega \to \reals^d$ and its corresponding diffusivity field $\kappa: [0,T] \times \Omega \to \reals_+$.
\footnote{See \cite{TD2002RW} for more details on the relation between the velocity and diffusivity fields.}
Then, the transport of a quantity of interest $c: [0,T] \times \Omega \to \reals$, e.g., a chemical concentration, in this domain is described by the advection-diffusion (AD) PDE \cite{NSADRE1996H}
\begin{equation} \label{eq:ADPDE}
\dotc = -\nabla \cdot( - \kappa \nabla c + \bbu \, c) + s ,
\end{equation}
where $\dotc = {\partial c}/{\partial t}$ denotes the time derivative of the concentration and $s: [0,T] \times \Omega \to \reals$ is the time-dependent source field.

Given an appropriate initial condition (IC) and a set of boundary conditions (BCs), it can be shown that the AD-PDE \eqref{eq:ADPDE} is well-posed and has a unique solution \cite{IFA1998R}. In this paper, we use the following IC and BCs
\footnote{Extending the results that follow to include more general BCs is straight-forward and is skipped here for simplicity of presentation.}:
\begin{subequations} \label{eq:IBCs}
\begin{align}
&c(0,\bbx) = g_0(\bbx) \ \ \, \for \bbx \in \Omega , \label{eq:IC} \\
&c(t,\bbx) = g_i(t,\bbx) \ \for \bbx \in \Gamma_i ,  \label{eq:BCs}
\end{align}
\end{subequations}
where $\Gamma_i$ for $i \in \set{1, \dots, n_b}$ denote the boundaries of $\Omega$ and $g_0: \Omega \to \reals$ and $g_i: [0,T] \times \Gamma_i \to \reals$ possess appropriate regularity conditions \cite{IFA1998R}.
Equation \eqref{eq:IC} describes the initial state of the field at time $t=0$ whereas equation \eqref{eq:BCs} prescribes the field value along the boundary $\Gamma_i$ as a time-varying function.

We refer to the parameters that appear in the AD-PDE \eqref{eq:ADPDE}, i.e., the velocity, diffusivity, and source fields as well as  the IC and BCs, as the input data. Any of these fields can depend on secondary parameters which consequently means that the solution of the AD-PDE will depend on those secondary parameters.
For instance, in source identification problems, a parametric solution of the AD-PDE is sought as the properties of the source term $s(t, \bbx)$ vary. This parametric solution is then used to solve a PDE-constrained optimization problem \cite{meJ1}.
In general, secondary parameters can also appear in MOR of PDEs. In the following, we denote by $\bbp \in \ccalP \subset \reals^m$ the set of $m$ secondary parameters that the PDE input-data depend on, where $\ccalP$ is the set of feasible values.

\begin{rem} [Alternative PDEs]
In this paper we focus on the AD-PDE as an important case-study that models a wide range of physical phenomena from mass transport to population dynamics \cite{RDEA2011K}. Nevertheless, the method developed in this paper applies to any other PDE that can be solved using the FE method.
\end{rem}

\subsection{Training Problem}
%
Discretization-based numerical methods, e.g., the FE method, can be used to approximate the solution of the AD-PDE \eqref{eq:ADPDE} given an appropriate set of boundary-initial conditions (BICs) \eqref{eq:IBCs}. As discussed earlier, these methods although very expressive, are computationally demanding to store and evaluate. Furthermore, it is not straightforward to obtain parametric solutions of PDEs using these methods. As a result, a wide range of MOR methods have been proposed to obtain such parametric solutions. These MOR methods typically rely on numerical methods to obtain the desired reduced models.
Utilizing neural networks (NNs) to solve PDEs, allows us to directly obtain differentiable parametric solutions of PDEs that are computationally more efficient to evaluate and require considerably less memory to store.

Let $\bbtheta \in \reals^n$ denote the weights and biases of the NN, a total of $n$ trainable parameters. Then the solution of the AD-PDE can be approximated by the nonlinear output function $f(t,\bbx; \bbp, \bbtheta)$ of this NN, where $f : [0,T] \times \Omega \to \reals$ maps the spatiotemporal coordinates $t$ and $\bbx$ to the scalar output field of the PDE, given a value for the secondary parameters $\bbp$. Note that the input of the NN consists of the coordinates $t$ and $\bbx$ as well as the parameters $\bbp$ and its dimension is $1+d+m$. See \cite{DLP2017C} for a detailed introduction to NNs. In the following, we drop the arguments $\bbp$ and $\bbtheta$ whenever they are not explicitly needed.

\begin{prob} [Training Problem] \label{problem}
Given the PDE input-data and the NN function $f(\cdot)$ with $n$ trainable parameters, find the optimal value $\bbtheta^*$ such that
\begin{equation} \label{eq:lossSuperV}
\bbtheta^* = \argmin \int_{\ccalP} \int_T \int_{\Omega} \abs{ c(t, \bbx; \bbp, \bbtheta) - f(t, \bbx; \bbp, \bbtheta) }^2 d\bbx \, dt \, d\bbp,
\end{equation}
where $c(\cdot)$ denotes the true solution of the PDE.
\end{prob}
As discussed in Section \ref{sec:NNintro}, Problem \ref{problem} can be solved in different ways. The simplest approach is to use supervised learning to obtain an approximate solution of the PDE using labeled data collected using numerical methods and a loss function defined by objective  \eqref{eq:lossSuperV}. In this approach, the NN acts merely as a memory cell and an interpolator to store the solution more efficiently than storing the values at all grid points.
Note that this approach does not alleviate the need for discretization-based numerical methods but instead it relies on them.
Alternative methods exist that train the NN in an unsupervised way. A typical approach is to consider a loss function defined by the PDE residual obtained when the approximate solution $f(\cdot)$ is substituted in \eqref{eq:ADPDE}. This approach has two problems. First, for the AD-PDE \eqref{eq:ADPDE} it requires evaluation of second order derivatives of $f(\cdot)$. Estimating a function from its higher order derivatives is inefficient since differentiation only retains slope information and is agnostic to translation. Second, training the differential form of the PDE amounts to learning a complicated field by only considering values at a limited set of points, i.e., a measure-zero set,
ignoring correlations in space-time. In the following, we propose a different approach that addresses these issues by defining a loss function that relies on the variational form of the PDE.

%% file: OT.tex
\subsection{Variational Loss Function} \label{sec:loss}
The goal of Problem \ref{problem} is to learn the parameters $\bbtheta$ of the NN so that for all values of $\bbp$, the function $f(\cdot)$ approximates the solution of the AD-PDE \eqref{eq:ADPDE} as well as possible. To capture this, we need to define a loss function $\ell: \reals^n \to \reals_+$ that reflects how well the function $f(\cdot)$ approximates the solution of \eqref{eq:ADPDE}. To do so, we rely on the variational form of the AD-PDE \eqref{eq:ADPDE}. Particularly, let $v: [0,T] \times \Omega \to \reals$ be an arbitrary compactly supported test function. Multiplying \eqref{eq:ADPDE} by $v(t, \bbx)$ and integrating over spatial and temporal coordinates we get
\begin{align*}
\int_0^T \int_{\Omega} v \left[ \dotc + \nabla \cdot( - \kappa \nabla c + \bbu \, c) - s \right] d\bbx \, dt = 0, \ \ \forall v.
\end{align*}
Performing integration by parts for $\dotc$ we have
\begin{align*}
 \int_0^T \int_{\Omega} \dotc \, v d\bbx \, dt &= \int_{\Omega} \left[ c(T) v(T) - c(0) v(0) \right] d\bbx \\
& - \int_0^T \int_{\Omega} c \, \dot{v} d\bbx \, dt = - \int_0^T \int_{\Omega} c \, \dot{v} d\bbx \, dt .
\end{align*}
Note that $v(0) = v(T) = 0$ since $v(t, \bbx)$ is compactly supported over the temporal coordinate. Similarly,
$$ \int_0^T \int_{\Omega} v \, \nabla \cdot \kappa \nabla c \, d\bbx \, dt = 
- \int_0^T \int_{\Omega} \nabla v \cdot \kappa \nabla c \, d\bbx \, dt , $$
where again the boundary terms vanish since $v(t, \bbx)$ is compactly supported.
Finally, note that for velocities far below the speed of sound the incompressibility assumption holds and $\nabla \cdot \bbu = 0$; see \cite{IFD2000B}. Putting together all of these pieces, we get the variational form of the AD-PDE as 
\begin{equation} \label{eq:varPDE}
l(c,v) = \int_0^T \int_{\Omega} \left[ \nabla c \cdot \left( \kappa \nabla v + \bbu \, v \right) - c \, \dot{v} - s \, v \right] \, d\bbx \, dt = 0.
\end{equation}

This variational form only requires the first-order spatial derivative and also an integration over a non-zero measure set as opposed to a single point. The test function acts as a weight on the PDE residual and the idea is that if \eqref{eq:varPDE} holds for a reasonable number of test functions $v(t, \bbx)$ with their compact supports located at different regions in space-time $[0,T] \times \Omega$, the function $f(\cdot)$ has to satisfy the PDE.
A very important feature of the test function $v(t, \bbx)$ is that it is compactly supported. This allows local treatment of the PDE as opposed to considering the whole space-time at once and is the basis of the FE method \cite{FEM2012H}. We discuss the explicit form of the test function $v(t, \bbx)$ in Appendix \ref{sec:testFun}.

Given the variational form \eqref{eq:varPDE}, we can now define the desired loss function. Consider a set of $n_v$ test functions $v_k(t, \bbx)$ sampling the space-time $[0,T] \times \Omega$, a set of $n_0$ points $\bbx_k \in \Omega$ corresponding to the IC, and sets of $n_{b,i}$ points $(t_k, \bbx_k) \in [0,T] \times \Gamma_i$ for the enforcement of the BCs. Then, for a given set of PDE input-data, specified by $\bbp$, we define the loss function $\ell: \reals^n \times \ccalP \to \reals_+$ as
\begin{align} \label{eq:loss}
\ell(\bbtheta, \bbp) &= w_1 \sum\nolimits_{k=1}^{n_v} \abs{ l(f, v_k) }^2  \nonumber \\
& + \frac{w_2}{n_0} \sum\nolimits_{k=1}^{n_0} \abs{f(0,\bbx_k) - g_0(\bbx_k)}^2  \nonumber \\
& + \frac{w_3}{\bbarn_b} \sum\nolimits_{i=1}^{n_b} \sum\nolimits_{k=1}^{n_{b,i}} \abs{f(t_k, \bbx_k) - g_i(t_k, \bbx_k)}^2 ,
\end{align}
where $l(f, v_k)$ is defined by \eqref{eq:varPDE}, $\bbw \in \reals_+^3$ stores the penalty weights corresponding to each term, and $\bbarn_b = \sum_{i=1}^{n_b} n_{b,i}$ is the total number of training points for the BCs. Note that in the first term in \eqref{eq:loss}, the integration is limited to the support of $v(t, \bbx)$ which is computationally very advantageous.
In the next terms, normalizing by $n_0 \and \bbarn_b$ makes the weights $\bbw$ independent of the number of training points. 

Next, consider a set of $n_p$ points $\bbp_j \in \ccalP$ sampling the space of the secondary parameters. Then, integrating \eqref{eq:loss} over the secondary parameters we obtain the desired total loss function $\ell: \reals^n \to \reals_+$ as
\begin{equation} \label{eq:lossTot}
\ell(\bbtheta) = \sum\nolimits_{j=1}^{n_p} \ell(\bbtheta, \bbp_j).
\end{equation}
This loss function is lower-bounded by zero. Since this bound is attainable for the exact solution of the AD-PDE \eqref{eq:ADPDE}, the value of the loss function is an indicator of how well the NN approximates the solution of the PDE.
Training using loss function \eqref{eq:lossTot} is an instance of unsupervised learning since the solution is not learned from labeled data. Instead, the training data here are unlabeled samples of space-time and the physics captured by the AD-PDE \eqref{eq:ADPDE} guides learning of the parameters $\bbtheta$. In that sense, our approach is a model-based method as opposed to a merely statistical method that automatically extracts features and cannot be easily interpreted.

\begin{rem} [Numerical Computation]
The derivatives that appear in the variational form \eqref{eq:varPDE} can be evaluated in a variety of ways. Here we use the automatic differentiation method available in \textsc{TensorFlow}. Furthermore, to evaluate the integral over the support of the test functions, we utilize the Gauss-Legendre quadrature rule as is common in the FE literature \cite{FEM2012H}. See Appendix \ref{sec:lossComp} for more details on the computation of the loss function \eqref{eq:lossTot}.
\end{rem}

\subsection{Optimal Training Points} \label{sec:AL}
As discussed in Section \ref{sec:NNintro}, existing NN methods to solve PDEs use arbitrary training points. 
Instead, here we optimally select the training points over the space-time to expedite the training process and increase the accuracy of the resulting solutions. This is very similar to the idea of local adaptive mesh refinement in the FE literature \cite{IA2005HCB}. Note that often the PDE solution varies more severely at certain regions of the space-time and adding more training points at those regions ensures that the NN adequately captures these variations.
In order to prioritize more informative training points, we employ the residual field of the PDE. For a given set of input data specified by $\bbp$, plugging the NN function $f(\cdot)$ into the PDE \eqref{eq:ADPDE}, we obtain the PDE residual field $r: [0,T] \times \Omega \to \reals$ as
\begin{equation} \label{eq:res}
r(t, \bbx) = \dot{f} + \nabla \cdot( - \kappa \nabla f + \bbu \, f) - s .
\end{equation}

Given this metric, we sample space-time $[0,T] \times \Omega$ using the rejection sampling algorithm \cite{IMCR2010RC} to generate training points that are proportionally more concentrated in regions with higher residual values. To this end, let $\hhatr(\cdot)$ denote the normalized metric defining a distribution over the space-time. Furthermore, let $g: [0,T] \times \Omega \to \alpha$ denote a uniform distribution over the space-time from which we draw the candidate training points where $\alpha$ is a constant. Also, let $M > 0$ be another constant such that $\hhatr(t, \bbx) \leq M g(t, \bbx)$ over the space-time. Then, for a given candidate point $(t, \bbx)$, the rejection sampling algorithm consists of drawing a random variable $u \sim \uniform[0,1]$ and accepting the candidate point if
$ u < { \hhatr(t, \bbx) }/{[ M g(t, \bbx) ]} . $
Define
$ M = \max_{ [0,T] \times \Omega } { \hhatr(t, \bbx) }/{ \alpha } .$
Then, the acceptance condition simplifies to
\begin{equation} \label{eq:rejSmp}
u < \frac{ r(t, \bbx) }{\max r(t, \bbx) } .
\end{equation}
Note that according to \eqref{eq:rejSmp}, candidate points that are closer to regions with higher PDE residual values have higher chances of being selected. We approximately determine $\max r(t, \bbx)$ by sampling the PDE residual over a uniform grid.

\begin{rem} [Boundary-Initial Conditions]
A similar procedure is utilized for optimal selection of training points for the BICs. Particularly, we replace metric \eqref{eq:res} with the corresponding least-square error values between the prescribed IC and BCs and the NN output as given by the second and third terms in the loss function \eqref{eq:loss}.
\end{rem}

\begin{rem} [Secondary Parameters]
We take into account the effect of the secondary parameters $\bbp$ in sampling from the space-time, by integrating the residual field $r(t,\bbx; \bbp)$ over them. Note also that a similar approach can be used to also optimally select the samples $\bbp_j$ of the secondary parameters.
\end{rem}

\subsection{VarNet Deep Learning Library} \label{sec:}
The proposed \varnet algorithm combines the loss function \eqref{eq:lossTot} with the optimal selection method for the training points to approximate the solution of AD-PDEs.
\begin{algorithm}[t]
\caption{VarNet Algorithm}
\label{alg:VarNet}
\begin{algorithmic}[1]

\REQUIRE Space-time domain and PDE input data;

\REQUIRE Widths of MLP-NN layers;

\REQUIRE Number of training points $n_v, n_0, n_{b,i}, \and n_p$;

\REQUIRE Number of training epochs $N$ and weights $\bbw$ in \eqref{eq:loss};

\STATE Generate uniform training points over $[0,T] \times \Omega \times \ccalP$;	\label{line:grid}

\FOR{$e=1:N$}		\label{line:epoch}

	\FOR{$j=1:n_p$} 	\label{line:MOR}

		\STATE Update trainable parameters $\bbtheta_{e,j}$ via an optimizer;		\label{line:opt}

	\ENDFOR
	
	\IF{loss value has converged}	\label{line:conv}
		
		\STATE Add optimal training points using \eqref{eq:rejSmp};	\label{line:optPoint}
		
	\ENDIF
		
\ENDFOR

\end{algorithmic}
\end{algorithm}
The algorithm begins by requiring the properties of the spatial domain $\Omega$ and the time horizon $T$ as well as the input data to the AD-PDE, i.e., the velocity and corresponding diffusivity fields, the source field, and the BICs, as a function of spatial and temporal coordinates and possibly secondary parameters $\bbp$.
These inputs define an AD-PDE instance that is implemented in the \varnet library using the \code{ADPDE(domain, diff, vel, source, tInterval, IC, BCs, MORvar)} class where depending on the dimension the spatial domain $\Omega$, \code{domain} is defined by an instance of \code{Domain1D} or \code{PolygonDomain2D} class, and \code{MORvar} is an instance of \code{MOR} class for parametric problems.
Next, the \varnet Algorithm \ref{alg:VarNet} requires the width of the layers of the multi-layer perceptron (MLP) NN, used to capture the solution of the AD-PDE, as well as the number of training points $n_v, n_0, n_{b,i}, \and n_p$. This information defines the training problem as an instance of \code{VarNet(layerWidth, discNum, bDiscNum, tDiscNum)} class.
Finally, the number of training epochs as well as the penalty weights for the terms appearing in the loss function \eqref{eq:loss} must be given. 

Given the number of training points, in line \ref{line:grid}, the algorithm generates a uniform grid over space-time, its boundaries, and possibly the space of secondary parameters $\ccalP$.
The training process begins in line \ref{line:epoch} where in each epoch the algorithm iterates through all training data. The loop over the samples $\bbp_j$ of the secondary parameters is performed in line \ref{line:MOR}. Training on a subset of data, corresponding to one value of $\bbp_j$ at a time, amounts to the idea of batch-optimization and computationally facilitates the inclusion of arbitrarily large number of samples of the secondary parameters by breaking down the training points into smaller batches \cite{DLP2017C}.
In line \ref{line:opt}, the algorithm performs the optimization iteration for $\bbp_j$ at epoch $e$, updating the parameters $\bbtheta_{e,j}$ of the NN for all training samples of the space-time and a given set of PDE input-data. A variety of optimization algorithms, included in \textsc{TensorFlow}, can be used for this purpose; see \cite{DLP2017C} for details.
In line \ref{line:conv}, the algorithm checks the convergence of the training process and in line \ref{line:optPoint} it adds new optimal training points over the space-time according to the procedure outlined in Section \ref{sec:AL}. The number of these new training points is given as a fraction $q \in [0,1]$ of the original numbers $n_v, n_0, \and n_{b,i}$.
The training process is performed in the member function \code{VarNet.train(epochNum, weight, frac, batchNum)}.

The \varnet library \cite{VarNet} implements Algorithm \ref{alg:VarNet} and contains additional functionalities including data parallelism and extensive tools for report generation during training and post-processing of the trained NNs. 
 In Appendix \ref{app:VarNet} we present more details on the implementation of Algorithm \ref{alg:VarNet}. For more details on the \varnet library \cite{VarNet} and its functionalities refer to the code documentation.

\begin{rem} [Convergence]
By the universal approximation theorem, for a large enough number of trainable parameters $n$, the NN can approximate any smooth function \cite{MFNUA1989HSW}. Given that the AD-PDE \eqref{eq:ADPDE} has a unique smooth solution for an appropriate set of input data, the NN should converge to this solution when $n \to \infty$; see \cite{DLASPDE2017SS} for details.
\end{rem}

\begin{rem} [Parallelization]
Referring to Algorithm \ref{alg:VarNet}, parallelization of the training process is trivial. We can choose the number of samples $n_v, n_0, n_{b,i}, \and n_p$ in accordance with the available computational resources and decompose and assign the summations to different processing units.
\end{rem}

%% file: sim.tex
In this section we present numerical experiments to demonstrate the performance of the \varnet Algorithm \ref{alg:VarNet}. Particularly, we study four different types of problems with an increasing order of difficulty:
(i) 1D and 2D time-dependent AD problems with analytical solutions that are used as benchmarks in the FE literature. For highly advective cases with large Peclet numbers,\footnote{Peclet number is a measure of the relative dominancy of advection over diffusion and is defined as Pe = $u \, \hhatl /\kappa$, where $\hhatl$ is a characteristic length of $\Omega$, $u$ is a characteristic velocity, and $\kappa$ is the average diffusivity of the medium.}
a boundary layer is formed whose prediction becomes very challenging for discretization-based numerical methods. The objective here is to demonstrate the capability of our algorithm to solve problems that require very fine grids when solved by these methods. These two problems have constant input-data which often is the only case that is studied in the relevant literature.
(ii) A 2D-AD problem with an analytical solution and non-constant input data. This is a fabricated problem where we find the source field that generates a desired concentration field given a set of input data. This is a standard method used to check the convergence of FE methods.
(iii) A 2D-AD problem with compactly supported sources. In this case, we cannot obtain an analytical solution for the AD-PDE and only an approximation from numerical methods, e.g., the FE method, is available. The objective here is to examine the effect of localized source terms.
(iv) A 2D-AD problem whose input data correspond to realistic mass transport phenomena in turbulent flows. In fluid dynamics applications often the input data are obtained for turbulent conditions and are non-constant. An analytical solution is not an option then. This is the most general case that one encounters in using the AD-PDE.


We solve all cases using the \varnet library \cite{VarNet} that was developed as part of this paper. This library is written in \textsc{Python} using the \textsc{TensorFlow} module; see Appendix \ref{app:VarNet} for details about its implementation and the code documentation \cite{VarNet} for more details on its capabilities. We use the ADAM stochastic optimization algorithm to train the NNs \cite{ADAM2014KB}. The \varnet Algorithm \ref{alg:VarNet} is run on a single \textsc{NVIDIA GeForce RTX 2080 Ti} processor for each problem instance. In each case, we use a multi-layer perceptron (MLP) to capture the solution of the AD-PDE and unless otherwise specified, we use a \code{sigmoid} activation function for all neurons. For a given uniform grid over the space-time, we compute the approximation error as
\begin{equation} \label{eq:err}
\text{err} = \frac{ \norm{\bbf - \bbc} }{ \norm{\bbc} } ,
\end{equation}
where the vectors $\bbf \and \bbc$ stack the outputs of the NN and the exact solution evaluated at the grid points, respectively. We use an analytical solution or the FE approximate solution to compute the vector $\bbc$.\footnote{When the reported error values are biased against the NN solution due to, e.g., error in analytical solution, we add an asterisk to the reported value.}
To generate the required FE models, we use an in-house code based on the \textsc{DiffPack} library \cite{CPDE2013L}.
We set the fraction of the optimal training points in \varnet Algorithm \ref{alg:VarNet} to $q = 0.5$ and use short-hand notation `1Dt' to denote a one-dimensional time-dependent AD-PDE and so on for other cases.
When reporting the number of training points, we may only give $n_v$. The number of training points for BICs is then deduced from $n_v$. For instance if $n_v = 300 \times 40 \times 20$ for a 2Dt problem, then the first entry refers to discretization of temporal coordinate and the last two entries refer to discretization of spatial coordinates and thus, $n_0 = 40 \times 20$ and $n_{b,i} = 300 \times \bbarn$, where $\bbarn = 20 \, l_i$ and $l_i$ is the length of boundary $i$.

\subsection{1Dt AD-PDE with Analytical Solution} \label{sec:1dt}
In this section we study the performance of the \varnet Algorithm \ref{alg:VarNet} for the benchmark problem presented in \cite{1DLADE2015M}. This problem is defined for $T = [0,2] \and \Omega = [-1,1]$ as
\begin{align} \label{eq:ADPDE-1Dt}
& \frac{\partial c}{\partial t} + u \frac{\partial c}{\partial x} = \kappa \frac{\partial^2 c}{\partial x^2} , 	\nonumber \\
& c(x,0) = - \sin (\pi x) ,	\nonumber \\
& c(-1,t) = c(1,t) = 0 .
\end{align}
Similar to \cite{1DLADE2015M}, we fix the velocity $u = 1$ and study the performance of our approach for two diffusivity values $\kappa = 0.1/\pi \and \kappa = 0.01/\pi$. For the latter value, the analytical solution becomes numerically unstable.

\begin{figure*}
        \centering
        \begin{subfigure}[b]{0.3\textwidth}
                \includegraphics[width=\textwidth]{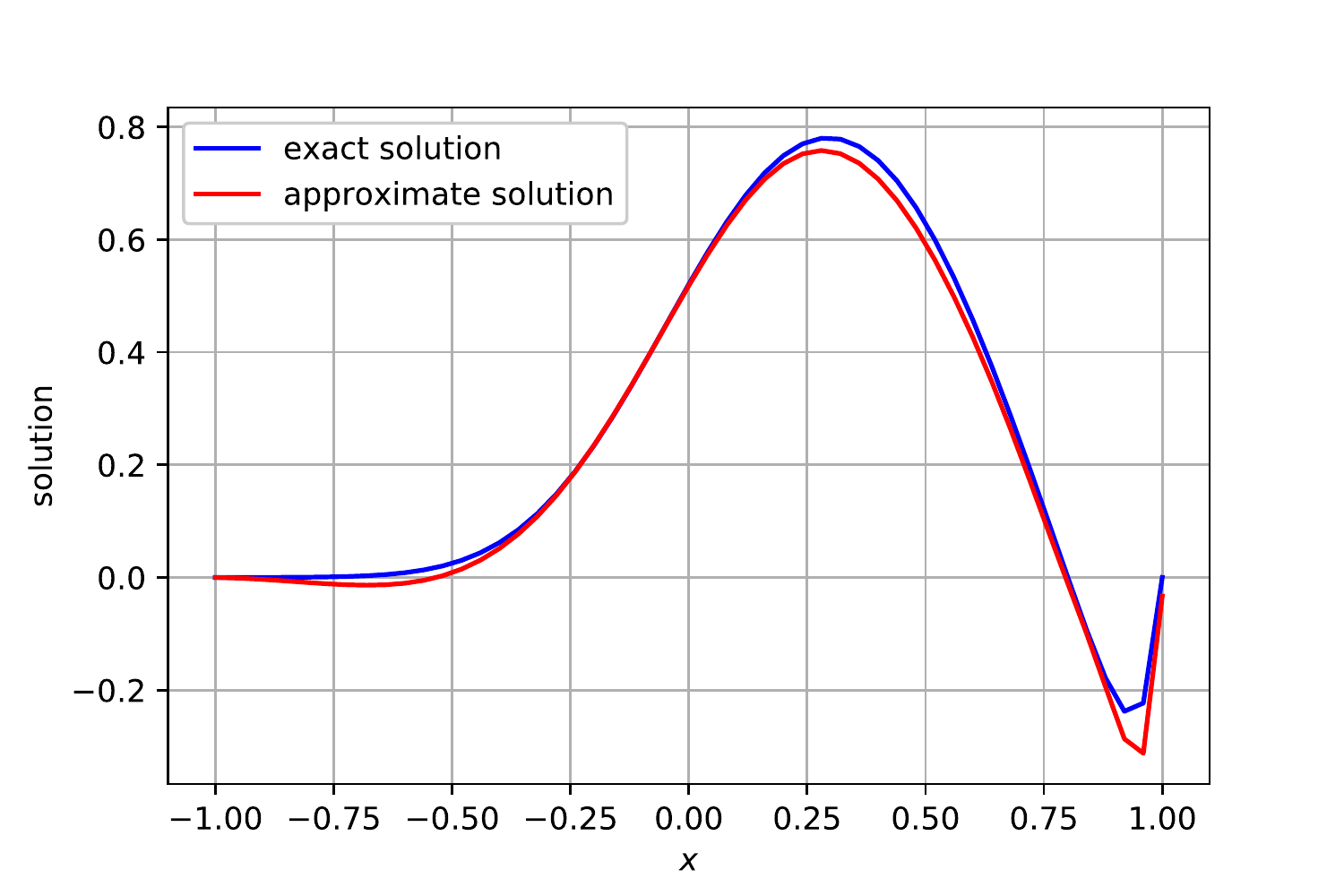}
    \caption{$t=0.8$}
    \end{subfigure}
    \begin{subfigure}[b]{0.3\textwidth}
                \includegraphics[width=\textwidth]{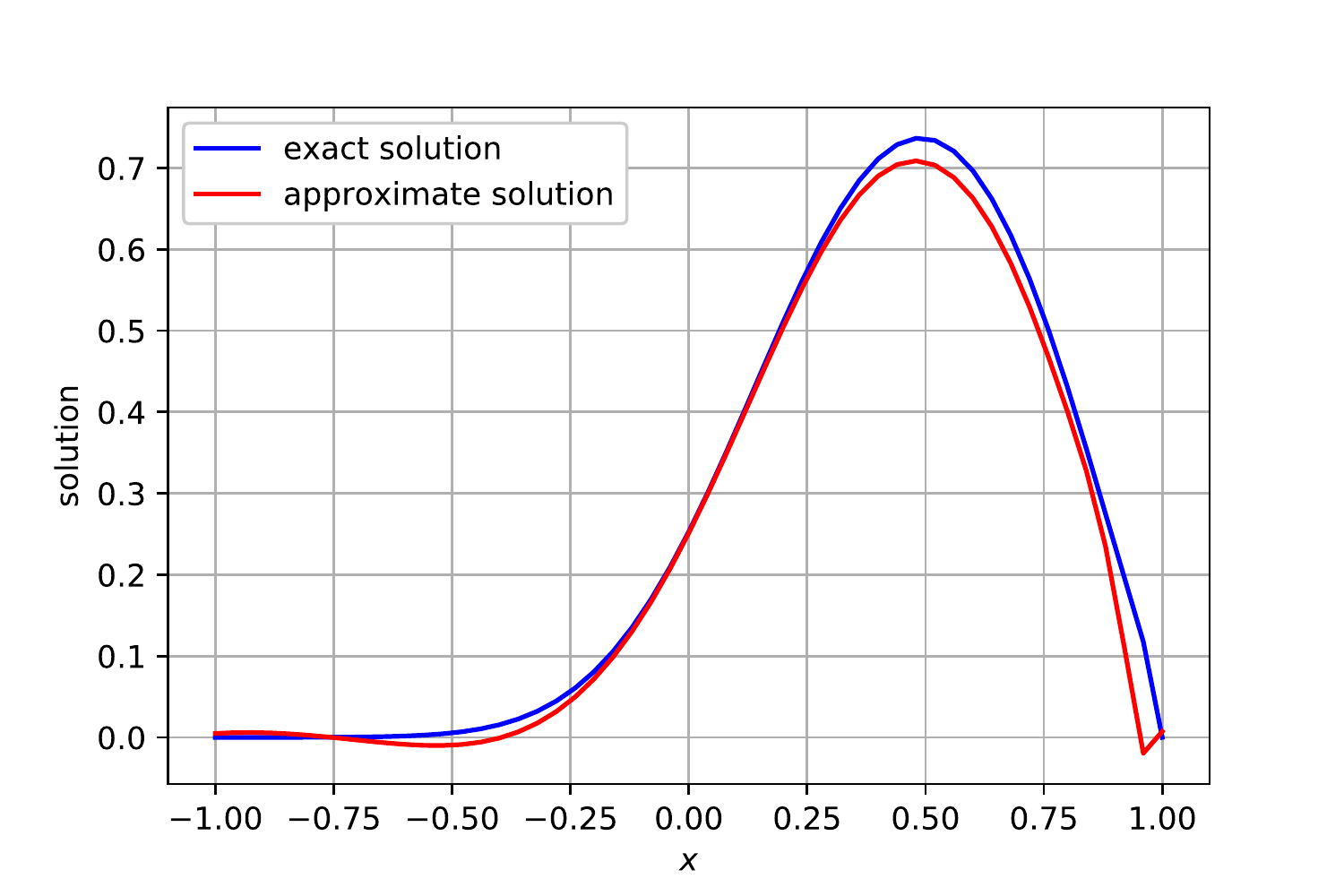}
    \caption{$t=1.0$}
\end{subfigure}
        \begin{subfigure}[b]{0.3\textwidth}
                \includegraphics[width=\textwidth]{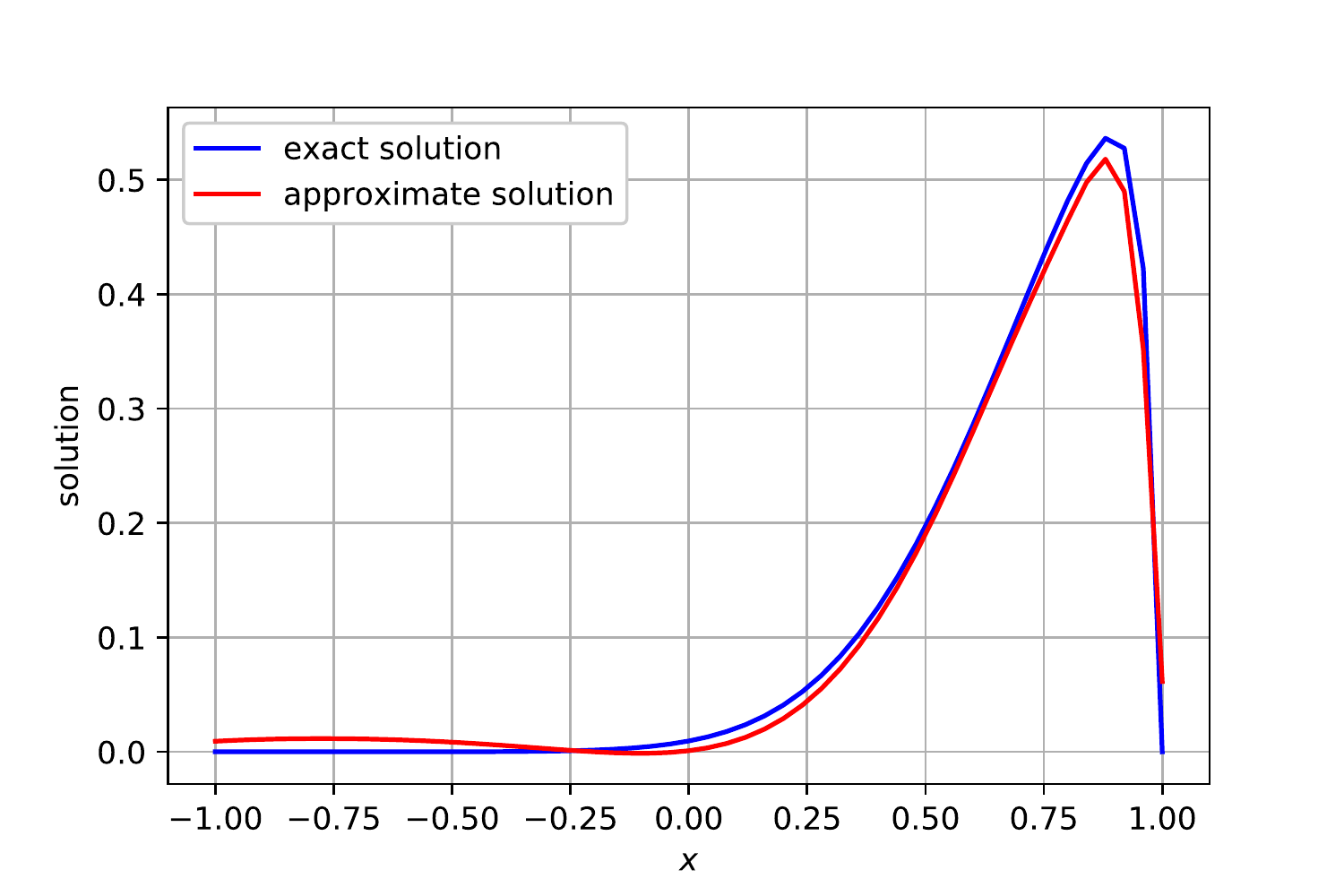}
    \caption{$t=1.6$}
    \end{subfigure}
\caption{Snapshots of the NN solution to 1Dt AD problem \eqref{eq:ADPDE-1Dt} for diffusivity $\kappa = 0.1/\pi$.}	\label{fig:1Dt-0_1sol}
\end{figure*}

Figure \ref{fig:1Dt-0_1sol} shows the solution, provided by the \varnet Algorithm \ref{alg:VarNet} for $\kappa = 0.1/\pi$ overlaid on the analytical solution. The final error \eqref{eq:err} is err $= 0.04$.
We use a single layer MLP with \code{sigmoid} activation, $20$ neurons in the layer, $n_v = 300 \times 20, n_0 = 20, \and n_{b,i} = 300$ training points, and set the weights to $\bbw = [1, 10, 10]$.
Note that the number of temporal training points should be sufficiently large to ensure that the dimensionless Courant number defined as
\begin{equation} \label{eq:courant}
C = u \frac{\Delta t }{ \Delta x},
\end{equation}
is upper-bounded and guarantee the stability of the solution across time \cite{FEMFP2003DH}. An animation of the approximate solution $f(t, x)$ for the case of $\kappa = 0.1/\pi$ is compared to the analytical solution in \cite{meJ5_video1}.  Note that the solution is continuous both in space and time and does not require interpolation to generate this video.

As mentioned earlier, one of the important contributions of this paper is the novel loss function that is based on the variational form of the AD-PDE as opposed to its differential form. To demonstrate the effectiveness of our loss function, we solve the case of $\kappa = 0.1/\pi$ using the PDE residual \eqref{eq:res} as well; see also the discussion pursuant to Problem \ref{problem}.
Using the same architecture and settings with $n_v = 1200 \times 80, n_0 = 80, \and n_{b,i} = 1200$ training points, which is equivalent to the number of integration points used above, the final error is err $= 0.88$. After a set of simulations, the best result corresponding to $n_v = 3000 \times 1000, n_0 = 1000, \and n_{b,i} = 3000$ and $\bbw = [1, 10, 5]$ has an error of err $= 0.50$. An animation of the approximate solution $f(t, x)$ for this case is compared to the analytical solution in \cite{meJ5_video3}. Observe that there is considerable error in capturing the solution and its BICs when the PDE residual is used, although the number of training points is considerably larger.

\begin{figure}[t]
    \centering
    \includegraphics[width=0.4\textwidth]{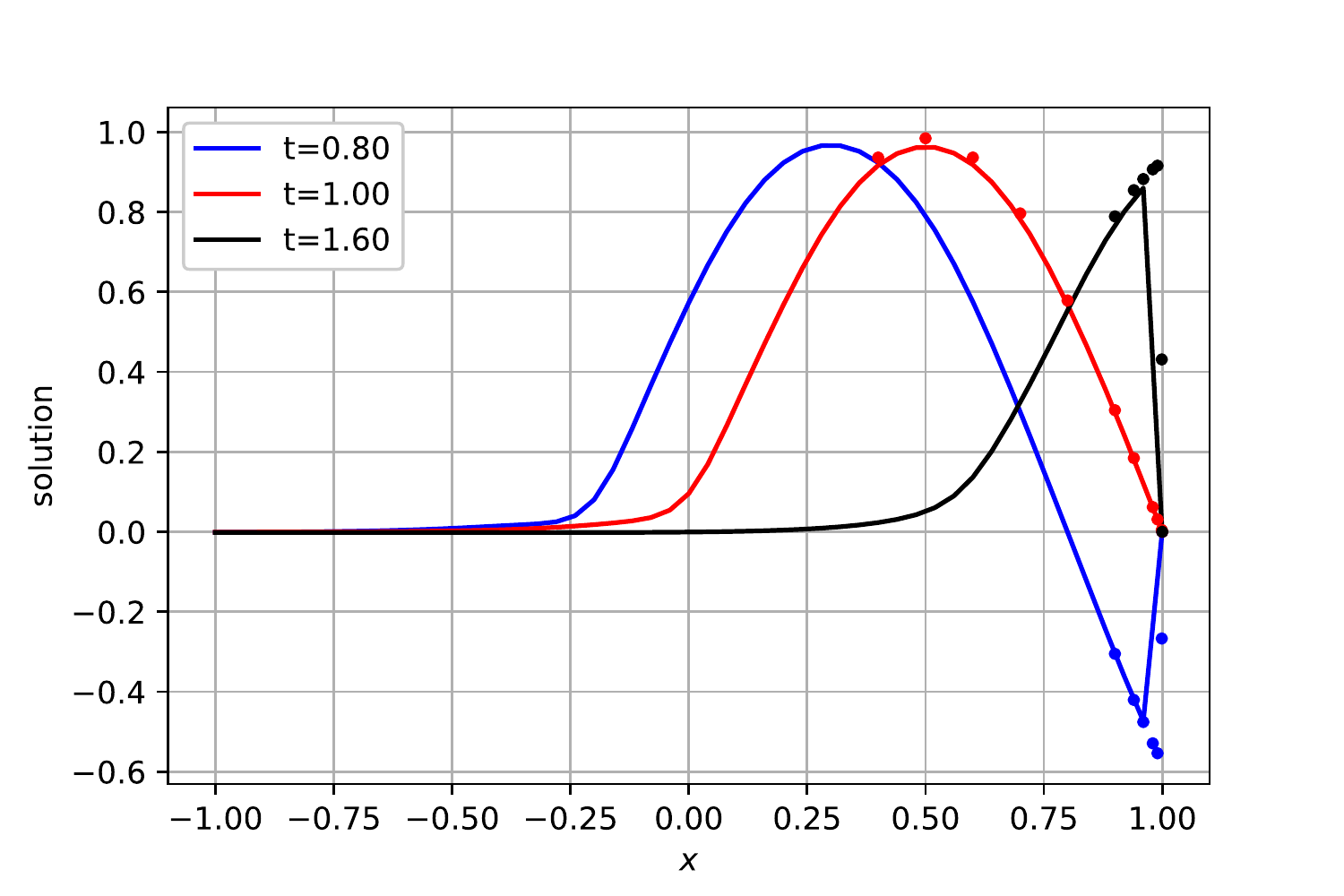}
\caption{Snapshots of the NN solution to 1Dt AD problem \eqref{eq:ADPDE-1Dt} for diffusivity $\kappa = 0.01/\pi$ overlaid on exact solution in select points shown by the dots.}	\label{fig:1Dt-0_01sol}
\end{figure}

Next, we consider the case of $\kappa = 0.01/\pi$ which corresponds to an order of magnitude higher Peclet number and is much more challenging to solve. Figure \ref{fig:1Dt-0_01sol} shows the snapshots of the solution from a MLP with two layers with $[10, 20]$ neurons in the layers amounting to $n=271$ trainable parameters. We train the NN using $n_v = 800 \times150, n_0 = 150, \and n_{b,i} = 800$ training points and set the weights to $\bbw = [1, 10, 10]$.
As mentioned earlier, for $\kappa = 0.01/\pi$ the analytical solution is numerically unstable and we only compare to a set of point values reported in \cite{1DLADE2015M}. The final error comparing to these point values is err* $= 0.09$. Note that this error value is conservative since it only considers the boundary layer, which is the most challenging to approximate, as opposed to the whole space-time.
Discretization-based methods have difficulty capturing the boundary layer at $x=1$ and require an extremely fine grid and consequently a large model \cite{1DLADE2015M}, whereas our approach can capture the solution with only $n = 271$ trainable parameters.
An animation of the approximate solution $f(t, x)$ for this case is given in \cite{meJ5_video2}.

\begin{figure}[t!]
        \centering
        \begin{subfigure}[b]{0.4\textwidth}
                \includegraphics[width=\textwidth]{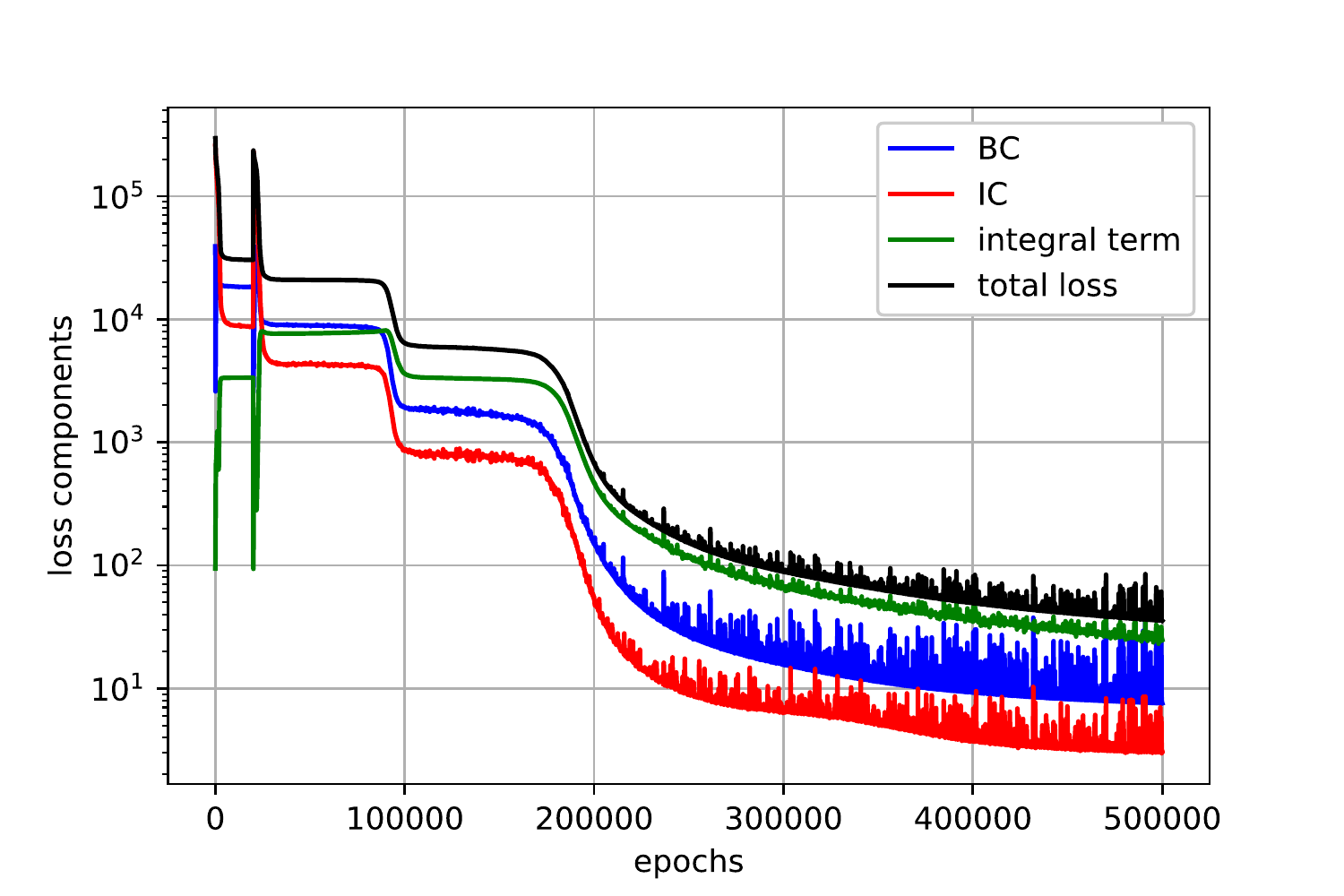}
    \caption{components and total value of loss function \eqref{eq:loss}}	\label{fig:1Dt-loss}
    \end{subfigure}
    \begin{subfigure}[b]{0.4\textwidth}
                \includegraphics[width=\textwidth]{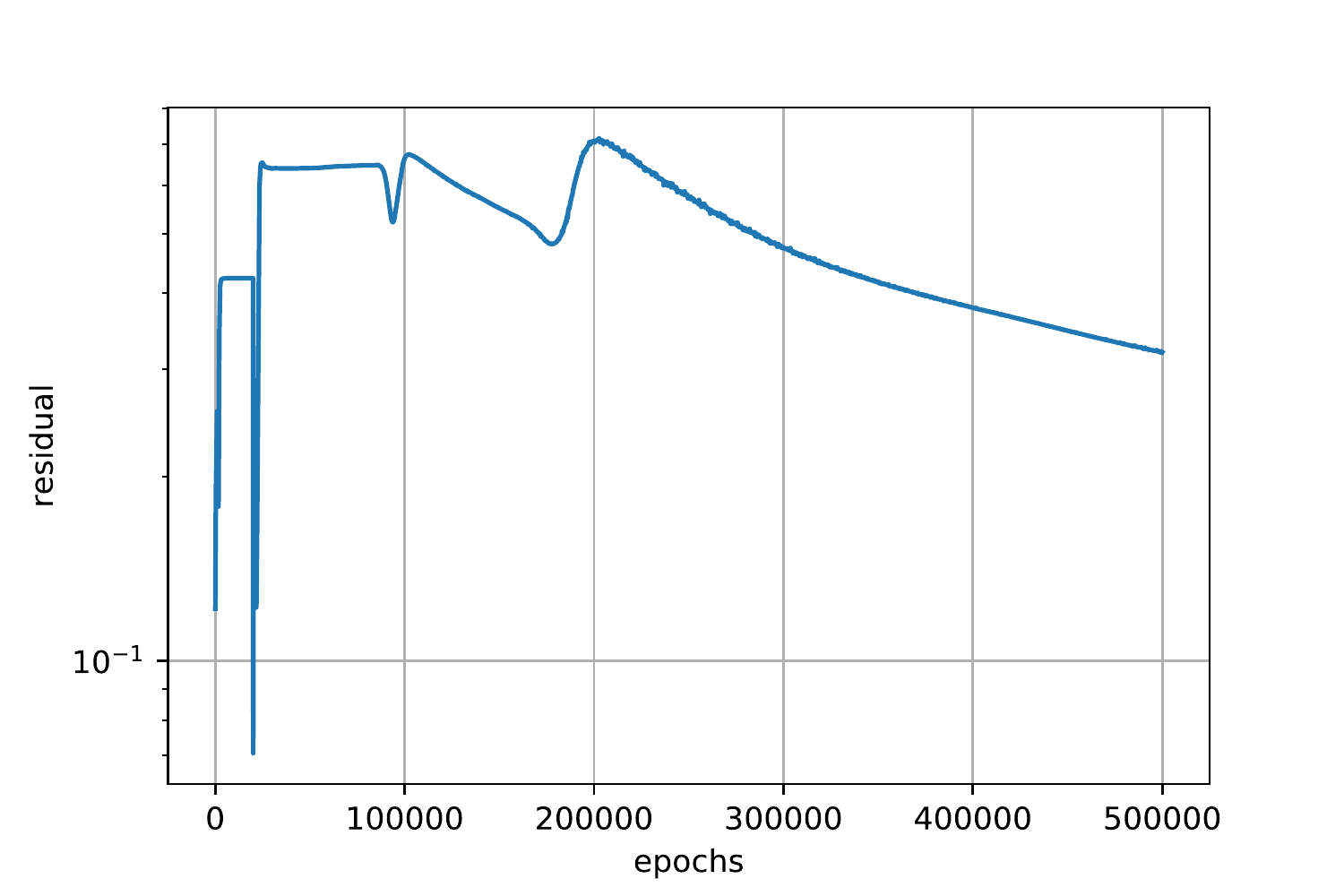}
    \caption{PDE residual \eqref{eq:res} convergence}	\label{fig:1Dt-resHist}
    \end{subfigure}
    \begin{subfigure}[b]{0.4\textwidth}
                \includegraphics[width=\textwidth]{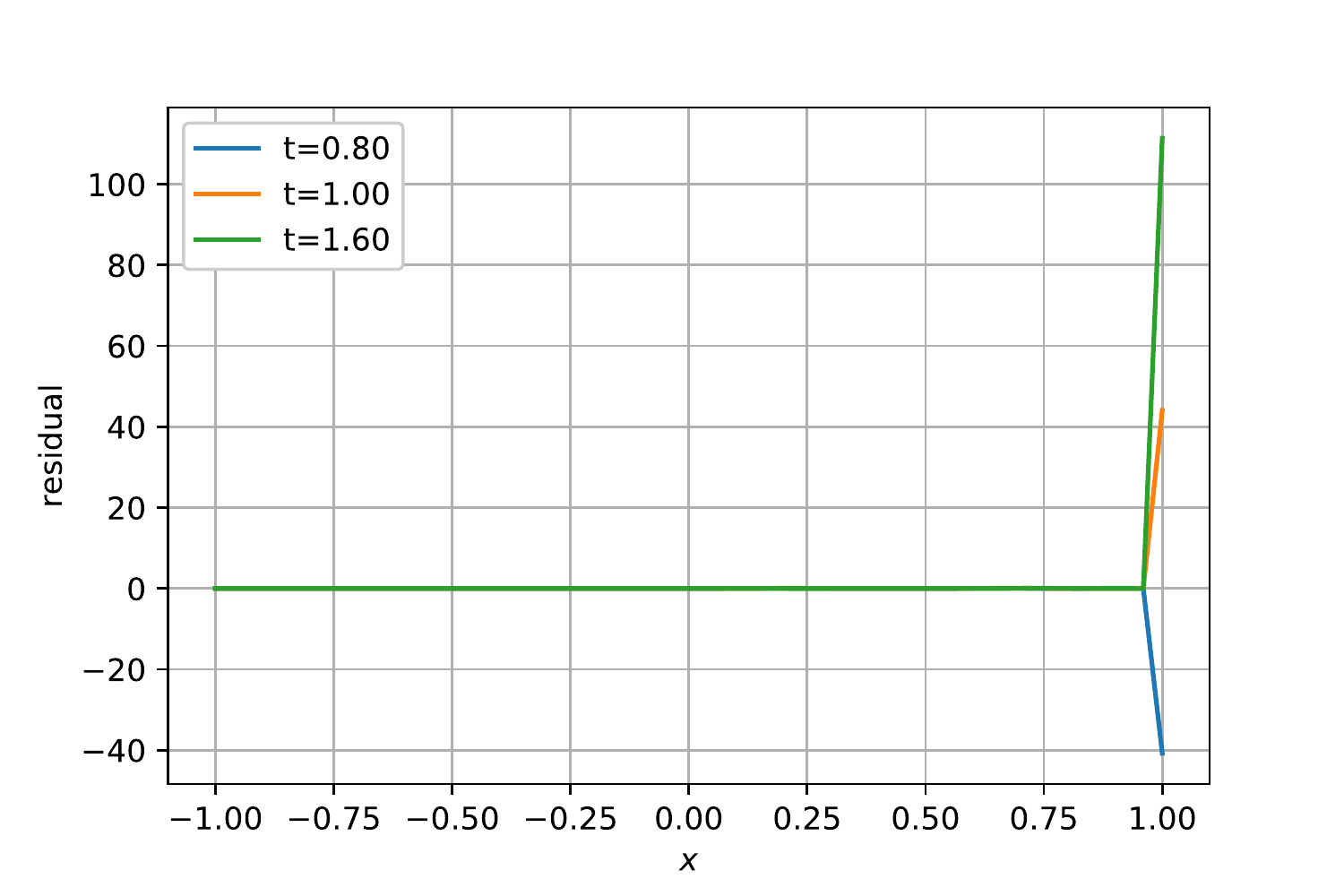}
    \caption{final residual snapshots}	\label{fig:1Dt-res}
    \end{subfigure}
\caption{Loss function \eqref{eq:loss}, PDE residual \eqref{eq:res} history, and final residual values for the 1Dt AD problem \eqref{eq:ADPDE-1Dt} with diffusivity $\kappa = 0.01/\pi$.}	\label{fig:1Dt-iterations}
\end{figure}

Figure \ref{fig:1Dt-iterations} shows the evolution of the individual terms and total value of the loss function \eqref{eq:loss}, the evolution of the PDE residual \eqref{eq:res}, and the corresponding final residual snapshots for the case of $\kappa = 0.01/\pi$.
Referring to Figure \ref{fig:1Dt-loss}, note that the penalty weights on the BICs in \eqref{eq:loss} must be selected as high as possible without making the variational term insignificant. Using small weights on BICs will result in inaccurate solutions since the AD-PDE does not have a unique solution without properly enforced BICs. For a properly selected set of weights $\bbw$, values of the BIC terms in the loss function \eqref{eq:loss} drop considerably fast to small values while the value of the variational term does not. This amounts to an increase in the PDE residual \eqref{eq:res} at the early training epochs, as can be seen from Figure \ref{fig:1Dt-resHist}. Once the BICs are approximately enforced, the contribution of the BIC terms in the loss function \eqref{eq:loss} decreases and the variational term starts to drop and the output of the NN converges to the unique solution of the 1Dt AD-PDE \eqref{eq:ADPDE-1Dt}; this amounts to a drop in the PDE residual at the later epochs; see Figure \ref{fig:1Dt-resHist}. The sharp increase around epoch $e = 2 \times 10^4$ corresponds to the addition of the optimal training points. Furthermore, from Figure \ref{fig:1Dt-res} it can be observed that the value of residual is considerably small across space-time except for $x=1$ which corresponds to the boundary layer.

\begin{figure}[t]
    \centering
    \includegraphics[width=0.4\textwidth]{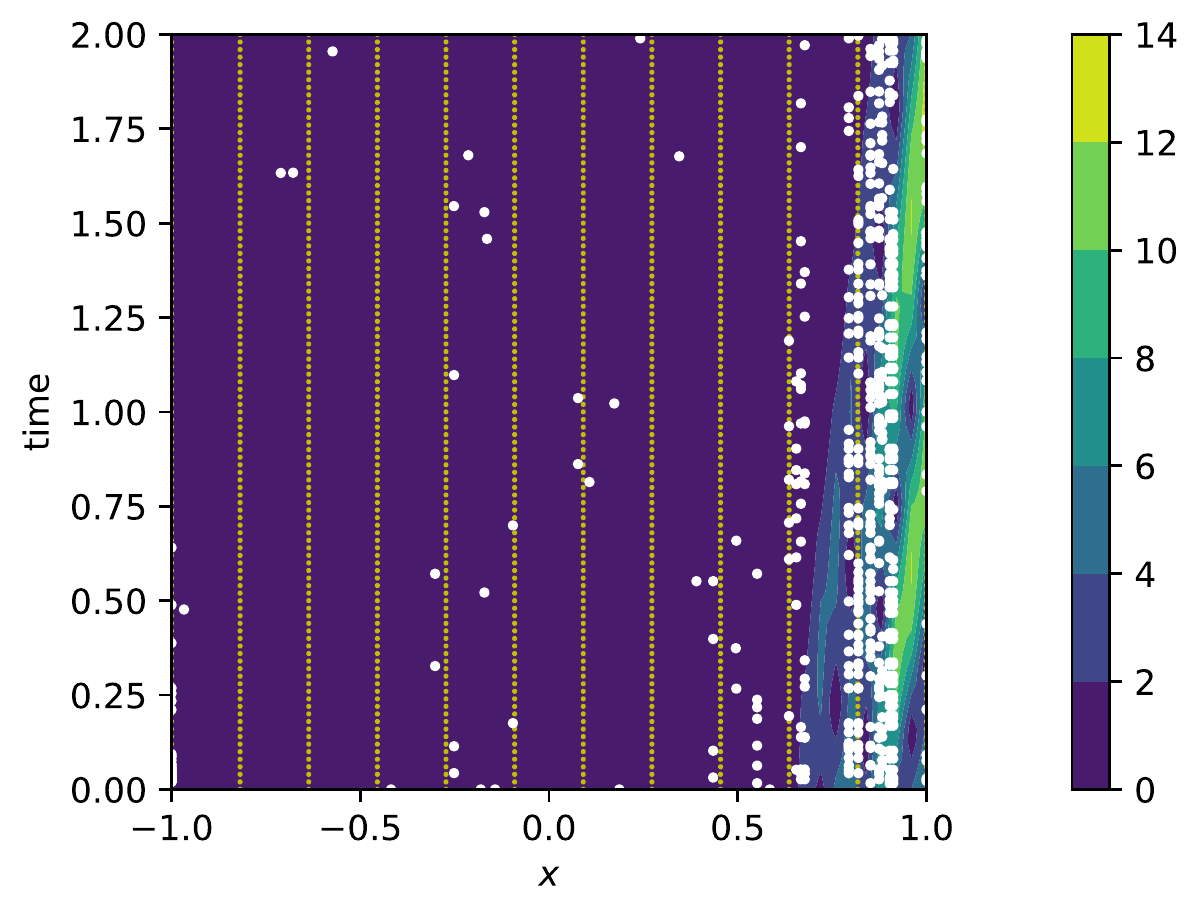}
    \caption{Training points for case 5 in Table \ref{table:err-1Dt} corresponding to the 1Dt AD problem \eqref{eq:ADPDE-1Dt} with $\kappa = 0.01/\pi$ overlaid on the PDE residual field \eqref{eq:res}. The green dots show the uniform training points across the space-time whereas the white dots show the optimally selected points according to the discussion of Section \ref{sec:AL}.}
    \label{fig:1Dt-optimPoints}
\end{figure}

Table \ref{table:err-1Dt} shows the performance of the \varnet Algorithm \ref{alg:VarNet}, in solving the AD problem \eqref{eq:ADPDE-1Dt} with diffusivity $\kappa = 0.01/\pi$, for different activation functions and network capacities where the number of training points and the weights are tuned for the best performance in each case.
\begin{table}[t!]
\centering
\renewcommand{\arraystretch}{1.25}
\caption{Error values \eqref{eq:err} for the 1Dt AD problem \eqref{eq:ADPDE-1Dt} with diffusivity $\kappa = 0.01/\pi$, calculated for the set of point values reported in \cite{1DLADE2015M}.}
\begin{tabular}{|c|c||c|c|c||c|} 
 \hline
No.	&	 $n$ 		& 	activation		&	$n_v$			& 	optimal 	&	err* 		\\ 	[0.5ex]  \hline\hline
1	&	 $81$	& \texttt{tanh}		&	$100 \times 10$	&	no		&	0.83 		\\ 	 \hline
2	&	 $81$	& \texttt{sigmoid}	&	$100 \times 10$	&	no		&	0.49 		\\ 	\hline \hline
3	&	 $81$	& \texttt{sigmoid}	&	$122 \times 13$	&	no		&	0.44 		\\ 	 \hline
4	&	 $81$	& \texttt{sigmoid}	&	$100 \times 10$	&	yes		&	0.42 		\\ 	 \hline
5	&	 $81$	& \texttt{sigmoid}	&	$100 \times 10$	&	scaled	&	0.39 		\\ 	 \hline \hline
6	&	 $271$	& \texttt{sigmoid}	&	$800 \times 150$	&	no		&	0.08 		\\ 	 \hline
7	&	 $271$	& \texttt{sigmoid}	&	$800 \times 150$	&	yes		&	0.09 		\\ 	 \hline \hline
8	&	 $911$	& \texttt{sigmoid}	&	$800 \times 150$	&	no		&	0.09 		\\ 	 \hline
9	&	 $911$	& \texttt{sigmoid}	& 	$1000 \times 200$	&	no		&	0.08		\\	 \hline
\end{tabular}
\label{table:err-1Dt}
\end{table}
Specifically, we examine three MLP structures with $[20], [10,20], \and [10,20,30]$ neurons in the layers. As mentioned before, $n_0$ and $n_{b,i}$ can be deduced from $n_v$ and are not reported.
Comparing the first two cases, observe that \texttt{tanh} activation is incapable of capturing the solution. This behavior is observed in other occasions as well and thus, in the rest of simulations we only use \texttt{sigmoid} activation.
The next three cases study the effect of using optimal training points, as proposed in Section \ref{sec:AL}, against uniform selection of the training points. More specifically, case 3 reports the error values for number of training points that is equivalent to addition of optimal points in cases 4 and 5. In case 5, the support of test functions used at optimal training points is scaled down by a factor of 10 at each dimension decreasing the error further. This is motivated by the argument in Section \ref{sec:AL} that the optimal training points should be placed in regions of high variability. Decreasing the support allows the NN to better capture the curvature at the boundary layer.
The white dots in Figure \ref{fig:1Dt-optimPoints} show the optimally selected training points for this case overlaid on the PDE residual field \eqref{eq:res}.
Note that as expected, most of the optimal training points are placed close to $x=1$ where the boundary layer is formed. This local refinement decreases the final error without having to uniformly increase the number of training points across space-time.
Further increasing the number of trainable parameters $n$ and the number of training points $n_v$ dramatically decreases the error as it can be seen in case 6. Note however that increasing $n$ without increasing $n_v$ unstabilizes the training process. Also the increase in the number of training points across temporal and spatial coordinates needs to be done proportionally to ensure that the Courant number \eqref{eq:courant} stays upper-bounded.
Comparing case 7 to case 6, observe that once the number of training points is increased enough so that the whole space-time is being adequately sampled, addition of optimal training points is not beneficial.
Finally considering cases 9 and 8, observe that further increase of the network capacity does not seem to decrease the error. Referring to Figure \ref{fig:1Dt-0_01sol} note that the error values reported in Table \ref{table:err-1Dt} effectively capture the error in boundary layer $x=1$. To capture this layer even better, the number of spatial samples and thus, to maintain the Courant number, the number of temporal samples need to be considerably increased.

\begin{figure}[t]
    \centering
    \includegraphics[width=0.4\textwidth]{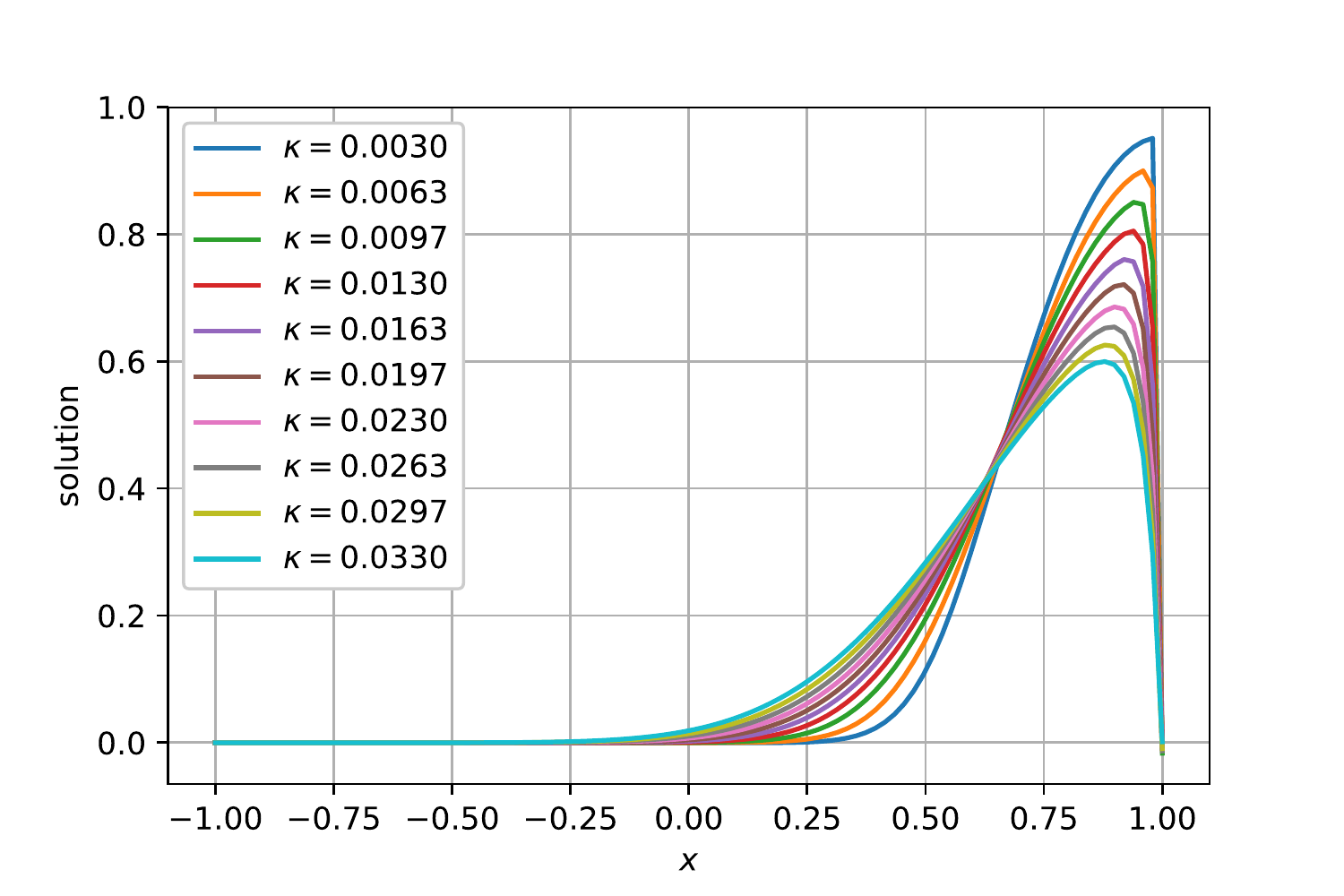}
    \caption{The parametric solution of the 1Dt AD problem \eqref{eq:ADPDE-1Dt} as a function of diffusivity $\kappa$ at $t=1.5$.}
    \label{fig:1Dt-MOR}
\end{figure}

Finally, we study the performance of the \varnet algorithm \ref{alg:VarNet} for MOR. Specifically, we train NNs that parametrically solve the AD-PDE \eqref{eq:ADPDE-1Dt} as a function of diffusivity for $\kappa \in [0.003, 0.033]$, which amounts to a range of more than one order of magnitude. We use $n_p=5$ geometric samples $\kappa \in \set{0.003, 0.0048, 0.0078, 0.0126, 0.0204, 0.033}$ for training. Table \ref{table:MORerr-1Dt} shows the final errors for diffusivity $\kappa$ values other than the ones used for training, and for different numbers of training points.
\begin{table}[t!]
\centering
\renewcommand{\arraystretch}{1.25}
\caption{Error values \eqref{eq:err} for the parametric solution of the 1Dt AD problem \eqref{eq:ADPDE-1Dt} as a function of diffusivity $\kappa$.}
\begin{tabular}{|c|c|c||c|c|} 
 \hline
No.	&	 $n$ 				&	$n_v$					& 	$\kappa$				&	err 		\\ 	[0.5ex]  \hline\hline
1	& \multirow{3}{*}{$101$}	& \multirow{3}{*}{$300 \times 20$}	& $0.01/\pi \approx 0.0032$	&	0.33* 	\\ 	 \cline{1-1} \cline{4-5}
2	&	 				&							& 	$0.005$				&	0.40*		\\ 	 \cline{1-1} \cline{4-5}
3	&	 				&							& $0.1/\pi \approx 0.03183$	&	0.14 		\\ 	 
\hline\hline
4	& \multirow{3}{*}{$921$}	& \multirow{3}{*}{$800 \times 150$}	& $0.01/\pi$				&	0.09* 	\\ 	 \cline{1-1} \cline{4-5}
5	&	 				&							& 	$0.005$				&	0.15*		\\ 	 \cline{1-1} \cline{4-5}
6	&	 				&							& $0.1/\pi$				&	0.00 		\\ 	 \hline
\end{tabular}
\label{table:MORerr-1Dt}
\end{table}
For the two smaller values of diffusivity, the analytical solution is unstable and only the point values reported in \cite{1DLADE2015M} are compared which results in biased (larger) error values for these cases.
Note that the parametric solutions are as good as the ones reported for individual diffusivity values above, meaning that MOR has no major effect in the solution predicted by the NNs. This is in contrast to traditional MOR methods that often introduce a large error in the estimation. Note also that the \varnet Algorithm \ref{alg:VarNet} is able to capture the solution of the AD-PDE \eqref{eq:ADPDE-1Dt} by only $n=921$ trainable parameters.
The continuous, parametric representation of the solution provides valuable tools for analysis of dynamical systems. For instance, we can study the sensitivity of the solution of the AD problem \eqref{eq:ADPDE-1Dt} to the change of diffusivity $\kappa$ of the medium. Figure \ref{fig:1Dt-MOR} shows the solution snapshots at $t = 1.5$ as a function of $\kappa$.
Note that as $\kappa$ decreases, the peak value is better preserved and carried downstream with less diffusion and as a result, the boundary layer becomes sharper.

\subsection{2Dt AD-PDE with Analytical Solution} \label{sec:2Dt_anal}
\begin{figure}[t]
    \centering
    \includegraphics[width=0.49\textwidth]{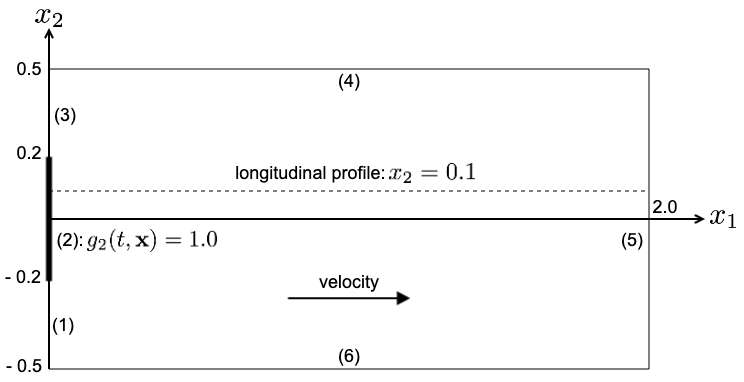}
    \caption{2D convex domain for 2Dt AD problem. The numbers in parentheses denote the index of the boundary segments. The constant concentration on the boundary segment (2) is transported downstream by advection and diffusion.}
    \label{fig:2Dt-domain}
\end{figure}

%
In this section we consider a 2D time-dependent AD problem with an analytical solution presented in \cite{ASADTDE1990LD}; this problem is used as a benchmark to evaluate the FE methods proposed in \cite{SADECDMFE1997SMAJ}. Specifically, a constant, one-directional flow with $\bbu = [1, 0]$ is considered and we adjust the constant diffusivity $\kappa$ to study the effect of the Peclet number on the solution.
We consider a convex 2D domain $\Omega = [0, 2] \times [-0.5, 0.5]$ where we assign a constant, non-zero Dirichlet condition to boundary 2 on the left side of the domain, i.e.,  $g_2(t,\bbx) = 1 \for t\in [0,T] \and \bbx \in \Gamma_2 = \set{\bbx \, | \, x_1 = 0 \and x_2 \in [-0.2, 0.2]}$; see Figure \ref{fig:2Dt-domain}.
In other words, the concentration on the boundary $\Gamma_2$ is transported downstream via advection and in the transverse direction via diffusion and at any given time $t$, a step-like concentration front exists at $x_1 = t$ that travels with a speed of $u_1=1$. At $t=0$ there is no concentration in the domain and we assume the boundaries are far enough so that they stay concentration-free for the whole time, i.e., in \eqref{eq:IBCs} we set $g_0(0, \bbx) = 0$ and $g_i(t, \bbx) = 0 \for i \in \set{1, 3, \dots, 6}$. Noting that the value of $T$ is immaterial as long as it is set large enough, we use $T = 1.5$.

\begin{figure*}
        \centering
        \begin{subfigure}[b]{0.3\textwidth}
                \includegraphics[width=\textwidth]{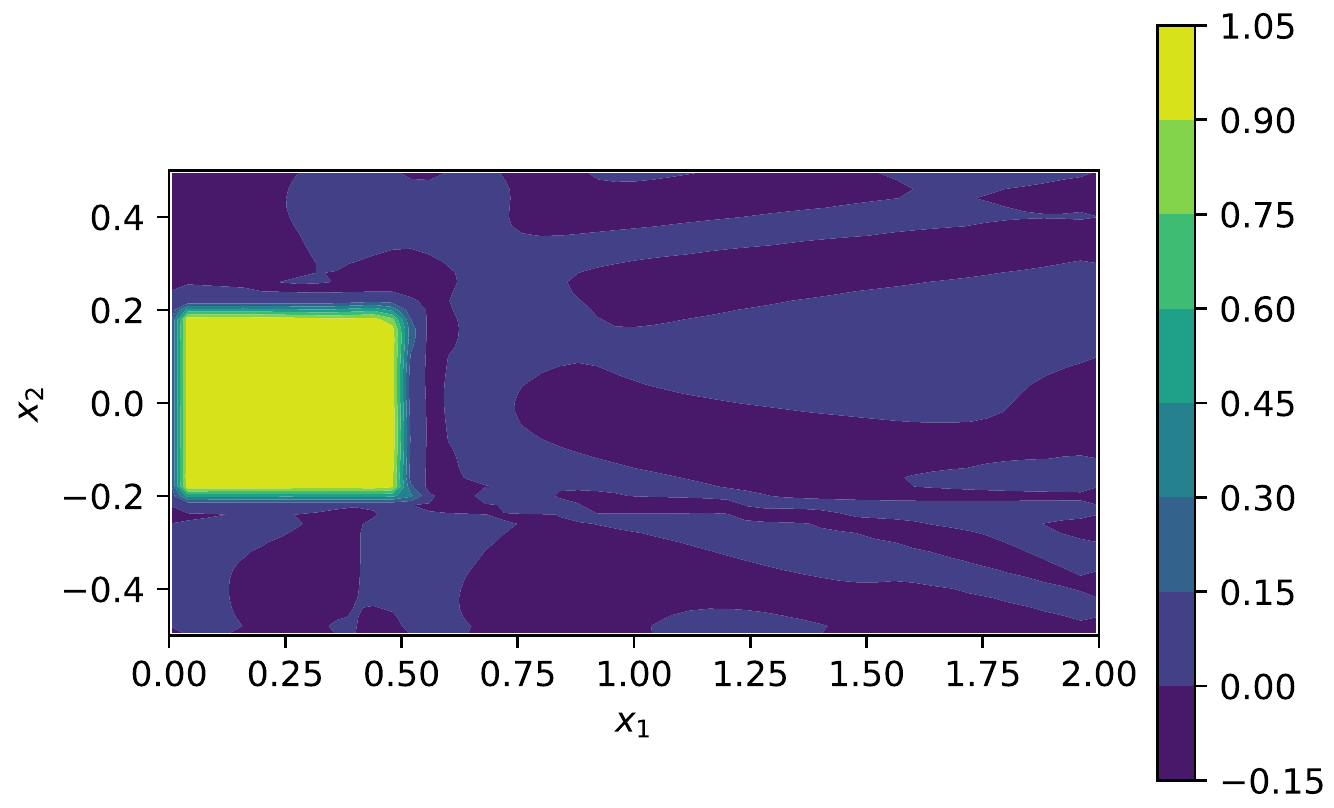}
    \caption{$t=0.5$}
    \end{subfigure}
    \begin{subfigure}[b]{0.3\textwidth}
                \includegraphics[width=\textwidth]{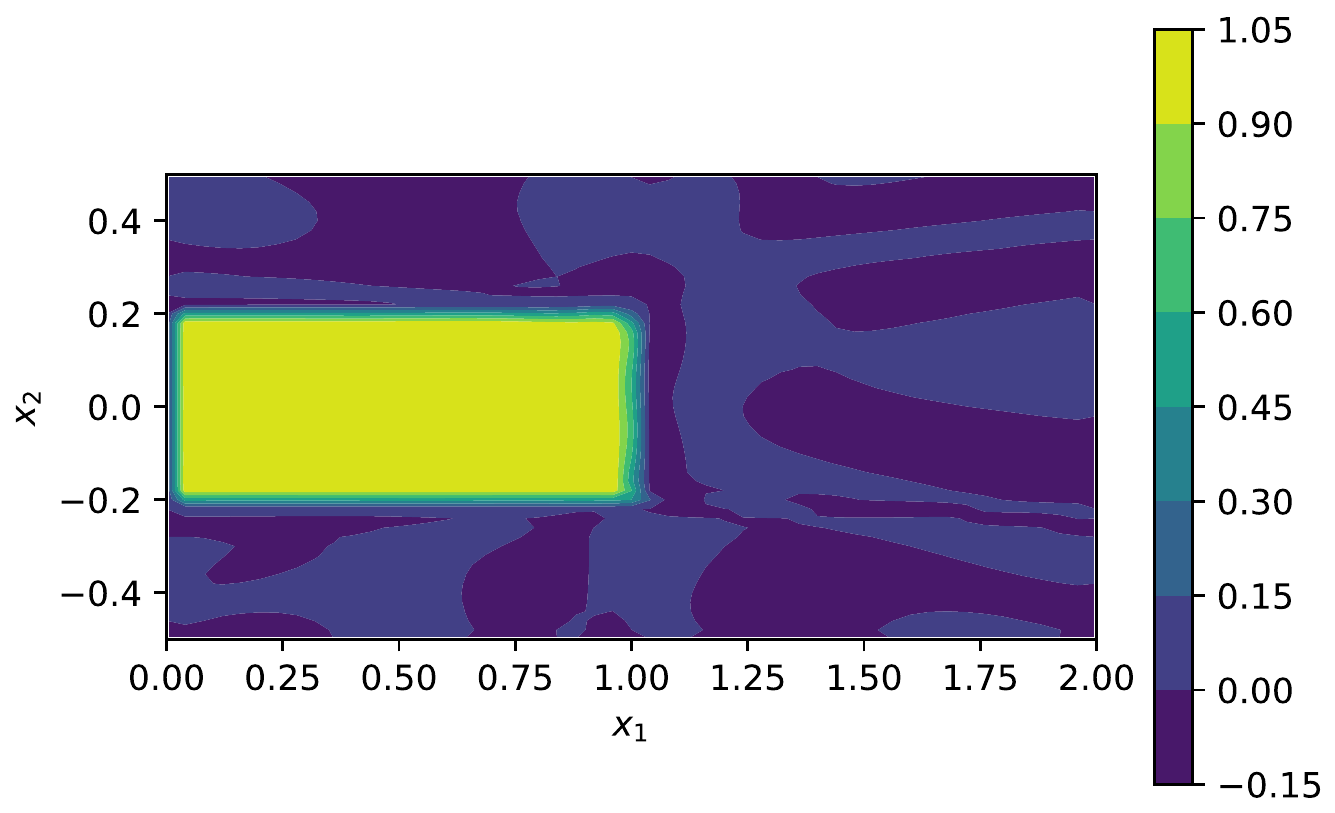}
    \caption{$t=1.0$}
\end{subfigure}
        \begin{subfigure}[b]{0.3\textwidth}
                \includegraphics[width=\textwidth]{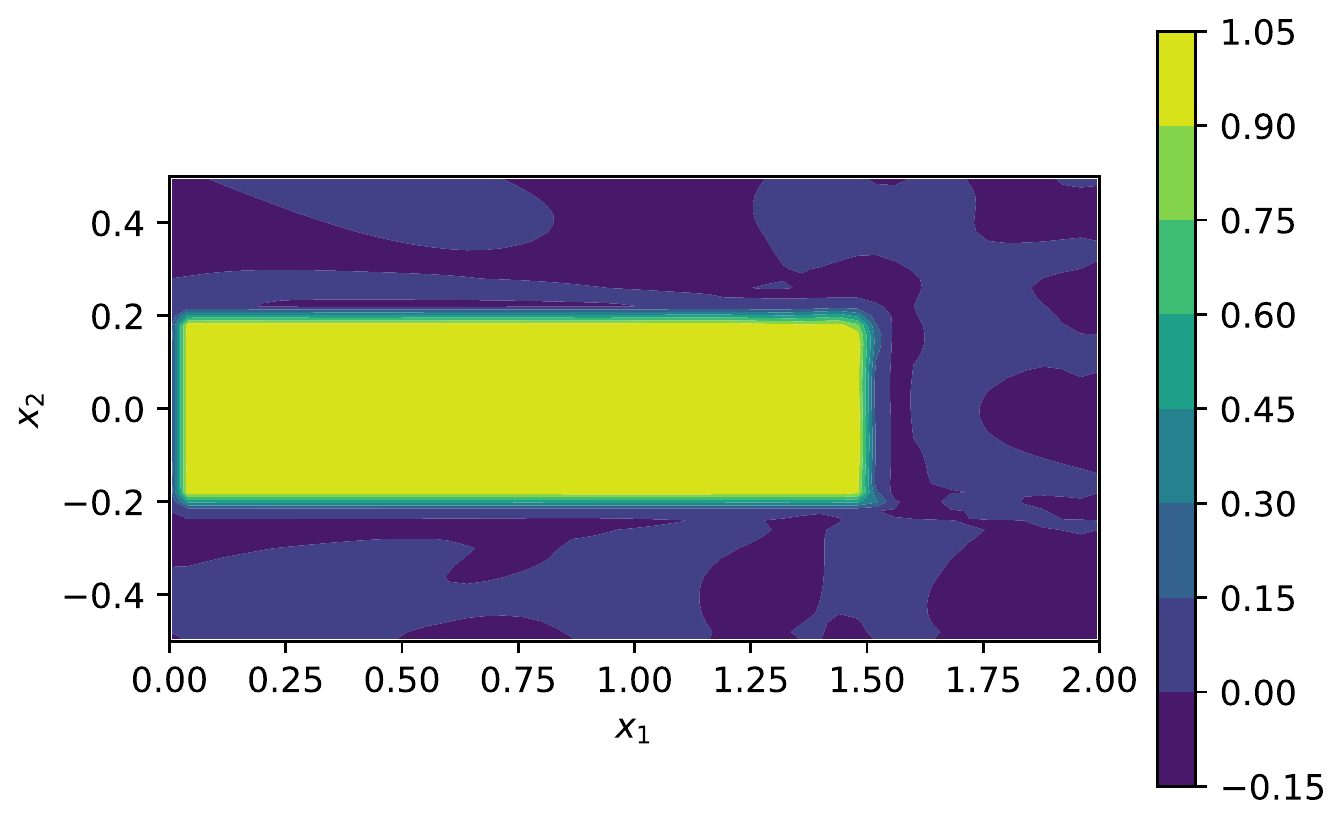}
    \caption{$t=1.5$}
    \end{subfigure}
\caption{Snapshots of the NN solution to the 2Dt AD problem for diffusivity $\kappa = 10^{-6}$.}	\label{fig:2Dt-snaps}
\end{figure*}

\begin{figure}[t]
    \centering
    \includegraphics[width=0.4\textwidth]{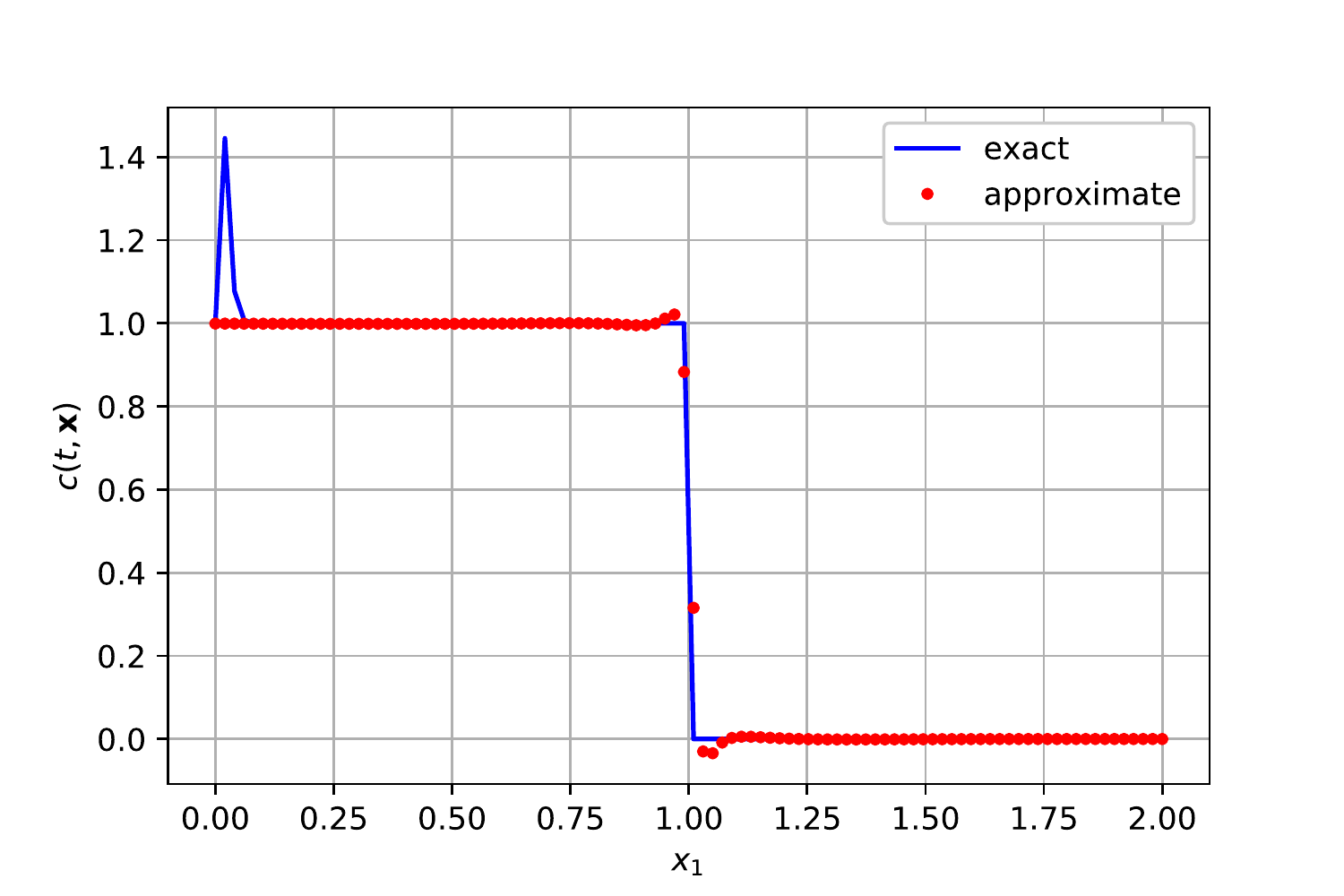}
    \caption{Exact and approximate concentration corresponding to the 2Dt AD-PDE along the longitudinal profile $x_2=0.1$ for $t = 1.0$.}
    \label{fig:2Dt-profile}
\end{figure}

In Table \ref{table:err-2Dt} we report the error values compared to the analytical solution provided in \cite{ASADTDE1990LD} for different diffusivity values where we set weights $\bbw = [1, 5, 1]$.
\begin{table}[t!]
\centering
\renewcommand{\arraystretch}{1.25}
\caption{Error values \eqref{eq:err} for the 2Dt AD problem for different constant diffusivity $\kappa$ values.}
\begin{tabular}{|c|c||c|c|c||c|} 
 \hline
No.	&	 $n$ 		& 	$\kappa$		&	$n_v$				& 	optimal 	&	err 		\\ [0.5ex] 
 \hline\hline
1	&	 $101$	& 	$10^{-3}$		&	$25 \times 20 \times 10$	& 	no		&	0.24 		\\ 	 \hline \hline
2	&	 $101$	& 	$10^{-3}$		&	$30 \times 23 \times 11$	& 	no		&	0.20 		\\ 	 \hline
3	&	 $101$	& 	$10^{-3}$		&	$25 \times 20 \times 10$	& 	yes		&	0.16 		\\ 	 \hline
4	&	 $101$	& 	$10^{-3}$		&	$25 \times 20 \times 10$	& 	scaled	&	0.18 		\\ 	 \hline \hline
5	&	 $101$	& 	$10^{-3}$		&	$75 \times 80 \times 40$	& 	no		&	0.15 		\\ 	 \hline
6	&	 $281$	& 	$10^{-3}$		&	$75 \times 80 \times 40$	& 	no		&	0.04 		\\ 	 \hline \hline
7	&	 $281$	& 	$10^{-6}$		& 	$100 \times 80 \times 40$	& 	no		&	0.07*		\\
 \hline
\end{tabular}
\label{table:err-2Dt}
\end{table}
Comparing the first 6 cases, observe that generally simultaneous addition of more training points $n_v$ and increasing of the capacity of the NN, i.e., the number of the trainable parameters $n$, lead to a decrease in the error value. Comparing cases 2, 3, and 4, it can be seen that using optimal training points results in a smaller error than a uniform grid with the same number of training points. Note that since in this problem, unlike the 1Dt AD problem \eqref{eq:ADPDE-1Dt}, the high variations are not concentrated at the boundary, scaling the support of optimal test functions in case 4 seems to have an adverse effect on the final error.
Case 7 shows the performance of the proposed method when the Peclet number is increased by three orders of magnitude. Note that this increase has no effect in the performance of the \varnet Algorithm \ref{alg:VarNet}. This is in contrast to FE methods which are known to become unstable for higher Peclet numbers \cite{SUPGFCDF1982BH}.
Figure \ref{fig:2Dt-snaps} shows the snapshots of the concentration field corresponding to case 7. Note the traveling step-like concentration front.
An animation of the solution filed $f(t, \bbx)$ can be found in \cite{meJ5_video4}. Figure \ref{fig:2Dt-profile} shows the exact and approximate concentration values along the longitudinal profile $x_2=0.1$ for $t = 1.0$; see also Figure \ref{fig:2Dt-domain}.
Note that the approximate concentration accurately matches the analytical solution of \cite{ASADTDE1990LD}. In fact the analytical solution itself is inaccurate close to boundary at $x_1=0$ meaning that the error err* $=0.07$ is biased in case 7 of Table \ref{table:err-2Dt}.

\subsection{2D AD-PDE with Analytical Solution and Field Input-Data}
The problems that we have studied so far have constant input data, i.e., the diffusivity and velocity fields as well as the BICs are constant and the source function is set to zero. Often, this is the only case studied in the relevant literature. When the PDE input-data are not constant, it is often impossible to obtain analytical solutions. In this section we construct a problem with non-constant input data and an analytical solution. Specifically, we start with concentration, diffusivity, and velocity fields, and use the AD-PDE \eqref{eq:ADPDE} to derive a corresponding source that will create this desired steady-state concentration field. Specifically, we consider a domain $\Omega = [-1,1]^2$ with the following concentration field
\begin{equation} \label{eq:concen_anal2D}
c(\bbx) = \sin(k \, \pi \, x_1) (1-x_2^2) ,
\end{equation}
where the integer $k \in \naturals_+$ adjusts the number of zero-crossing of the trigonometric function in the field. The larger $k$, the more complicated the resulting field is. Plugging this field into the AD-PDE \eqref{eq:ADPDE}, we obtain the corresponding source field $s(\bbx)$ that will generate the concentration field \eqref{eq:concen_anal2D} for a given set of PDE input-data. Note that this concentration field attains a value of zero on the boundaries so we use zero-valued Dirichlet BCs in \eqref{eq:BCs}. Since the problem is in steady state, the IC \eqref{eq:IC} is irrelevant here.

\begin{figure} [h!]
        \centering
	\begin{subfigure}[b]{0.38\textwidth}
		\includegraphics[width=\textwidth]{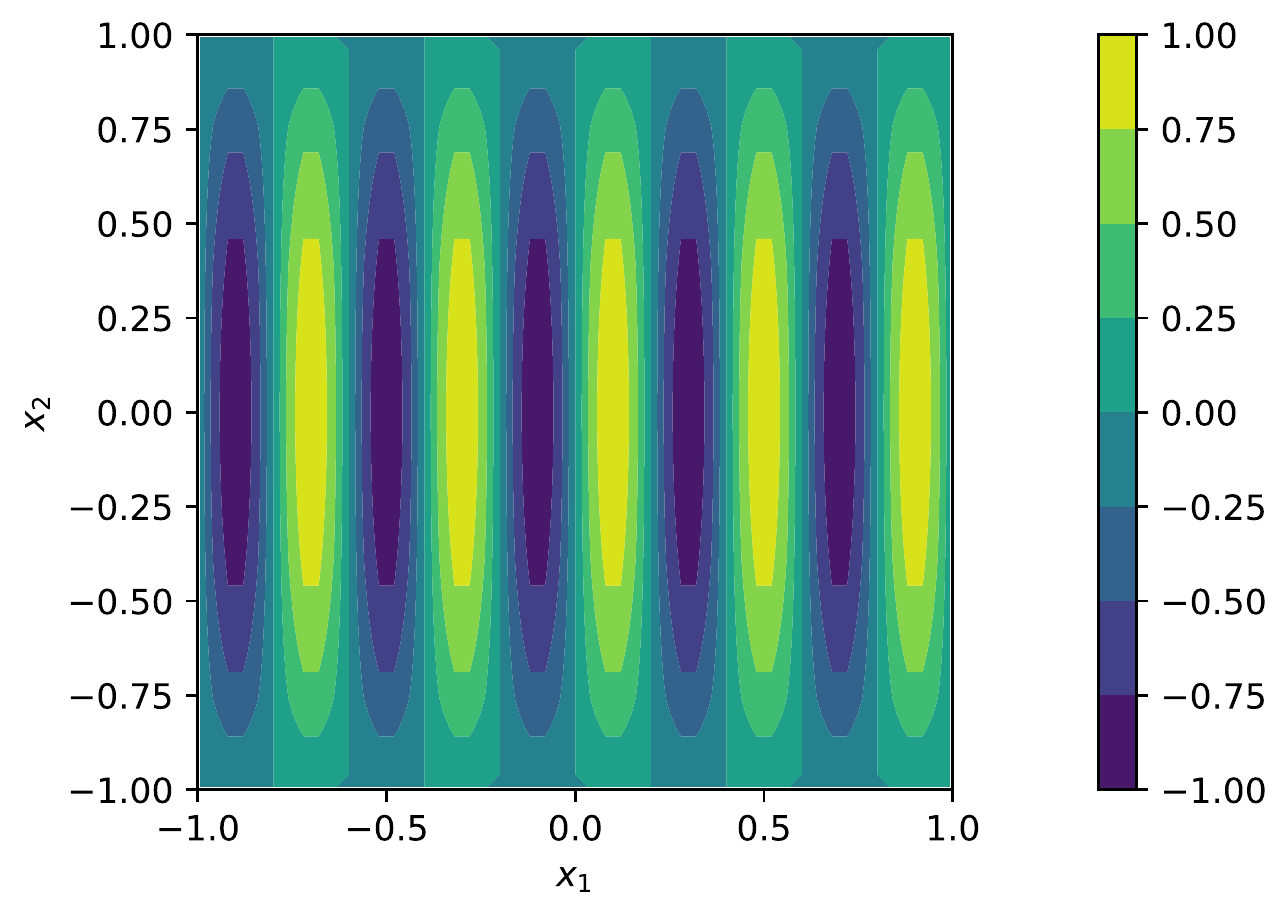}
	\caption{exact concentration field \eqref{eq:concen_anal2D}}		\label{fig:concen_anal2D}
	\end{subfigure}
	\begin{subfigure}[b]{0.38\textwidth}
		\includegraphics[width=\textwidth]{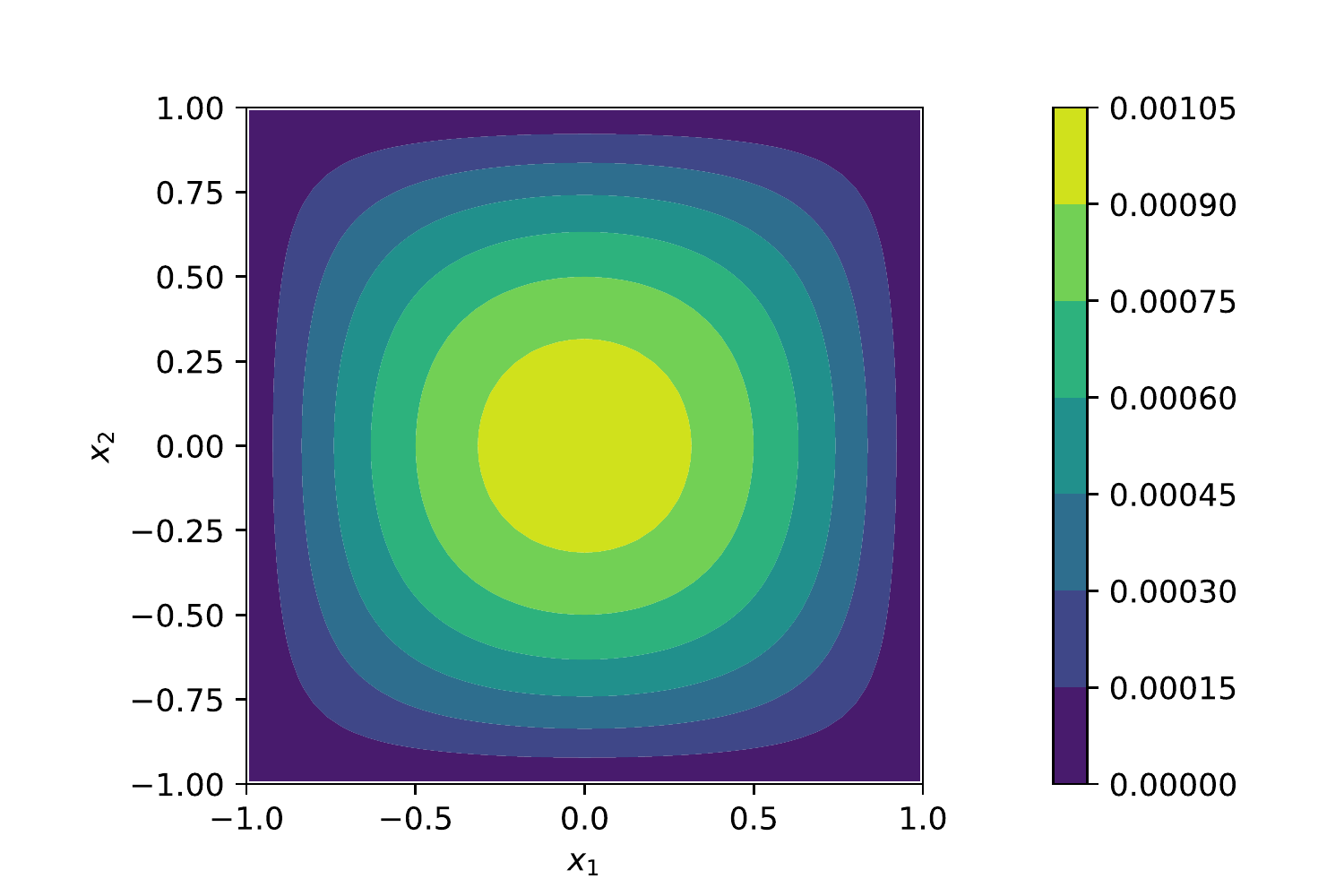}
	\caption{diffusivity field \eqref{eq:diff_anal2D}}		\label{fig:diff_anal2D}
	\end{subfigure}
	\begin{subfigure}[b]{0.38\textwidth}
		\includegraphics[width=\textwidth]{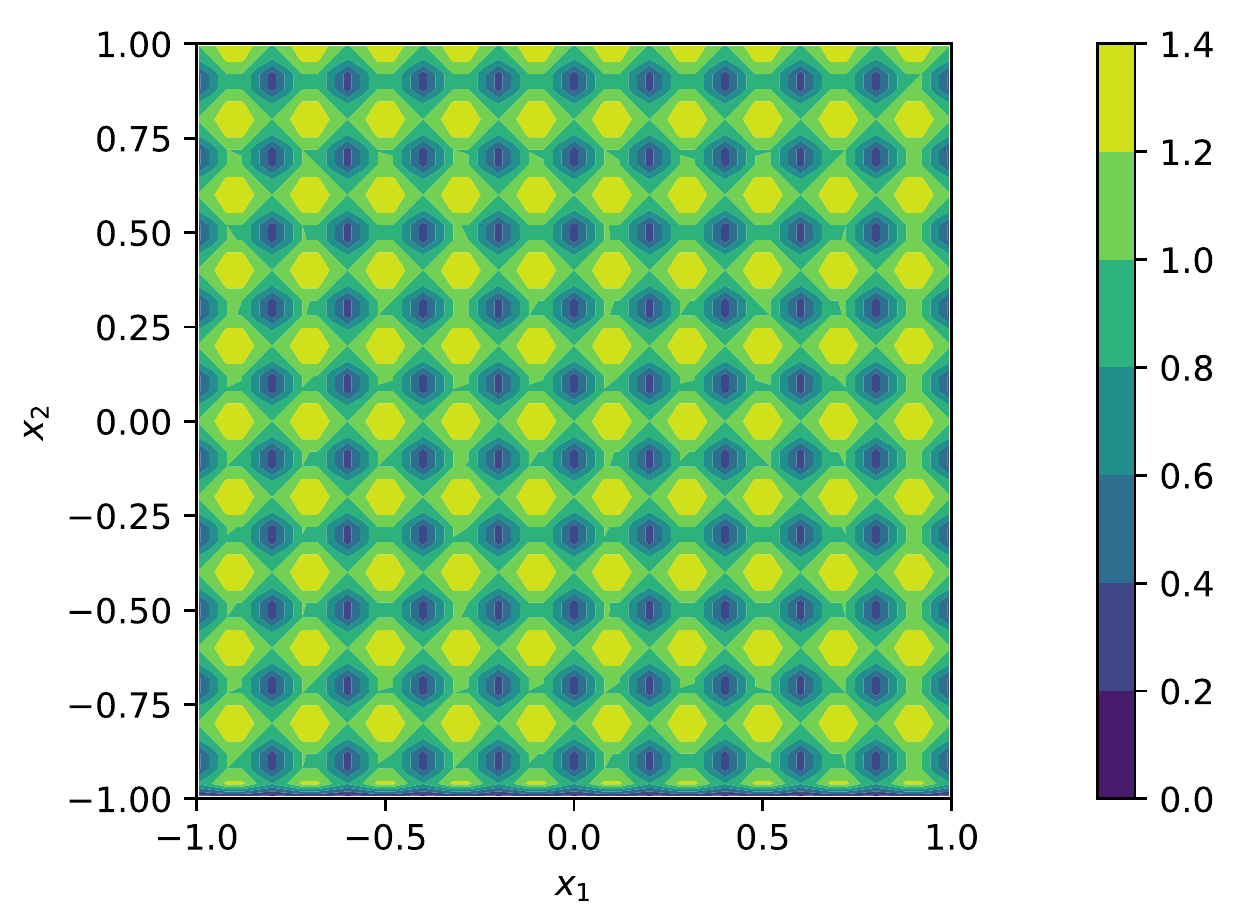}
	\caption{velocity magnitude field \eqref{eq:vel_anal2D}}		\label{fig:vel_anal2D}
	\end{subfigure}
	\begin{subfigure}[b]{0.38\textwidth}
		\includegraphics[width=\textwidth]{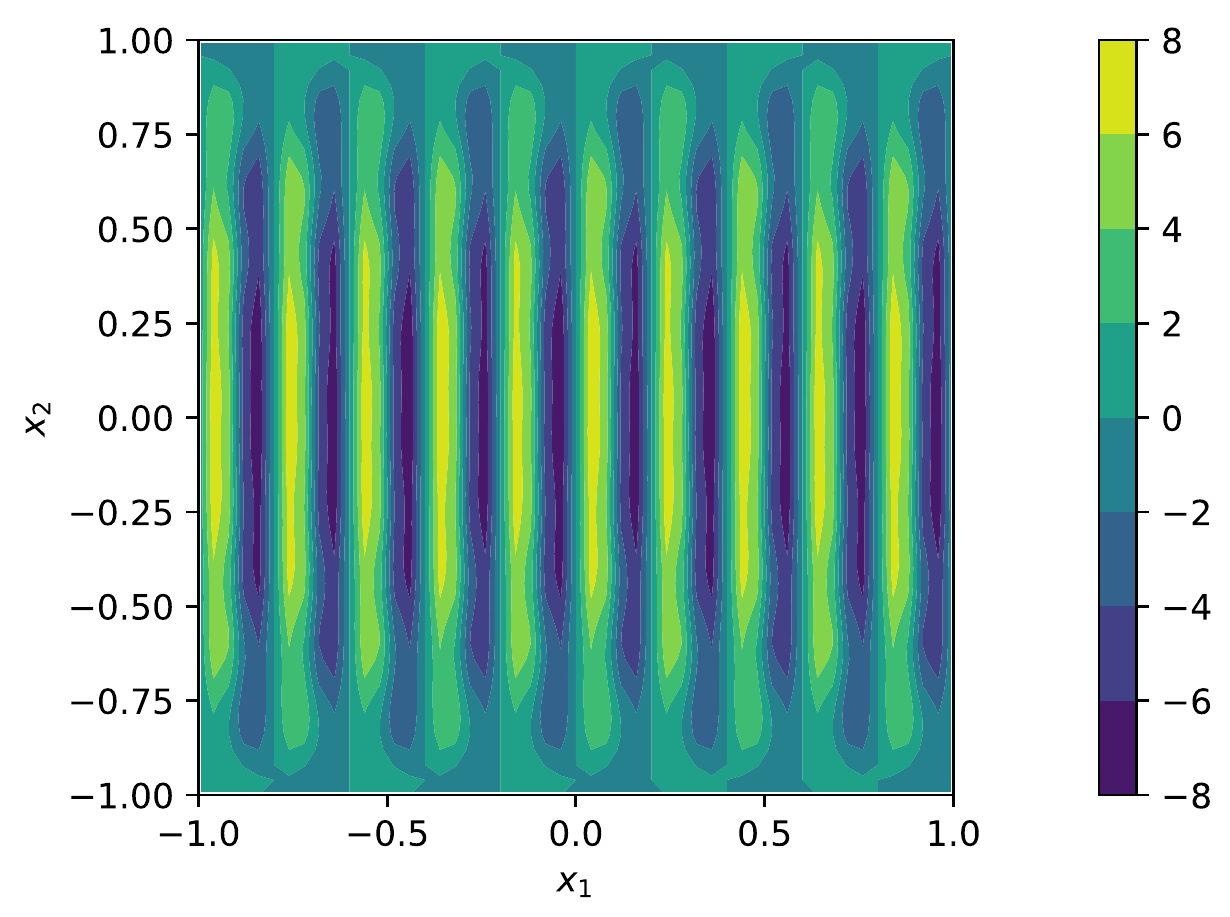}
	\caption{source field}		\label{fig:source_anal2D}
	\end{subfigure}
\caption{PDE input-data for the 2D AD problem with analytical fields. Specifically, these fields correspond to case 6 in Table \ref{table:err-anal2D}.}
\end{figure}

In Table \ref{table:err-anal2D}, we report the performance of the \varnet Algorithm \ref{alg:VarNet} in solving this problem instance where we set the weights $\bbw = [10, 1]$ for the reported results.
\begin{table}[t!]
\centering
\renewcommand{\arraystretch}{1.25}
\caption{Error values \eqref{eq:err} for the 2D steady-state AD problem with analytical solution \eqref{eq:concen_anal2D}.}
\begin{tabular}{|c|c||c|c|c|c||c|} 
 \hline
No.	&	 $n$ 		&	 $k$ 		& 	$\kappa$				& 	$\bbu$				&	$n_v$		&	err 		\\ [0.5ex] 
 \hline\hline
1	&	 $81$	&	 $1$		& 	$1$					& 	$[1, 0]$				&	$20 \times 20$	&	0.03 		\\ 	 \hline
2	&	 $81$	&	 $1$		& 	$1$					& 	$[1, 0]$				&	$40 \times 40$	&	0.04	 		\\ 	 \hline \hline
3	&	 $81$	&	 $1$		& 	$10^{-2}$				& 	$[1, 0]$				&	$40 \times 40$	&	0.03	 		\\ 	 \hline
4	&	 $81$	&	 $1$		& 	$10^{-3}$				& 	$[1, 0]$				&	$40 \times 40$	&	0.00	 		\\ 	 \hline \hline
5	&	 $81$	&	 $5$		& 	$10^{-3}$				& field \eqref{eq:vel_anal2D}	&	$40 \times 40$	&	0.04	 		\\ 	 \hline
6	&	 $81$	&	 $5$		& field \eqref{eq:diff_anal2D}	& field \eqref{eq:vel_anal2D}	&	$40 \times 40$	&	0.03	 		\\ 	 \hline
\end{tabular}
\label{table:err-anal2D}
\end{table}
Note that since the AD PDEs studied here and the following sections are time-independent, $w_2$ in \eqref{eq:loss} is trivially set to zero and is not reported. In the first four cases in Table \ref{table:err-anal2D}, $k=1$ and we use constant diffusivity and velocity fields, as reported in columns four and five. A single layer NN with 20 neurons can easily capture the concentration field \eqref{eq:concen_anal2D} in these cases regardless of the Peclet number. Comparing the first two cases, observe that increasing the number of training points $n_v$ does not improve the error. The reason could be that the current network capacity $n$ is insufficient to better capture the concentration field \eqref{eq:concen_anal2D}.
In cases 5 and 6, we use $k=5$ which results in the concentration field depicted in Figure \ref{fig:concen_anal2D}.
More specifically, in case 5 we use a non-constant velocity field and in case 6 we also use a non-constant diffusivity field. These two fields are defined as
\begin{subequations} \label{eq:2Danal-inpData}
\begin{align}
\kappa(\bbx) &= \bbarkappa \, (1-x_1^2) (1-x_2^2) ,	\label{eq:diff_anal2D} \\
\bbu(\bbx) &= \left[ \sin(k \, \pi \, x_1), \ \cos(k \, \pi \, x_1) \right] ,	\label{eq:vel_anal2D}
\end{align}
\end{subequations}
where $\bbarkappa \in \reals_+$ is the maximum diffusivity value in the domain $\Omega$ which we set to $\bbarkappa = 10^{-3}$. These two fields are depicted in Figures \ref{fig:diff_anal2D} and \ref{fig:vel_anal2D}, respectively. Figure \ref{fig:source_anal2D} shows the resulting source field corresponding to case 6.
The \varnet Algorithm \ref{alg:VarNet} successfully solves these two cases using only a single layer with $n=81$ trainable parameters.

\subsection{2D AD-PDE with Compactly Supported Source} \label{sec:compact}
The source field used in the previous section is not compactly supported; see Figure \ref{fig:source_anal2D}. Often in practice, the source term $s(t, \bbx)$ in AD-PDE \eqref{eq:ADPDE} is compactly supported and occupies a very small part of the domain. Existence of such source terms translates to a more localized and nonlinear concentration field that is more challenging to capture. To investigate this point further, in this section we solve the AD-PDE \eqref{eq:ADPDE} over a convex domain $\Omega = [-1,1]^2$ with constant input date $\kappa = 1$, $\bbu = [1, 0]$, and a set of Dirichlet BCs for two localized sources. Specifically, we first consider a Gaussian source given by
$$ s(\bbx) = A \exp \left( - \frac{ \norm{\bbx - \bbbeta}^2 }{ 2 \gamma^2 } \right) ,$$
where $A \in \reals_+$ is the source intensity, $\bbbeta = [-0.30, 0]$ is the source center and $\gamma$ is its characteristic length. This source technically is not compactly supported but for practical purposes it decays to almost zero beyond $\pm 3 \gamma$ distance from the center $\bbbeta$. Because of its smoothness, the corresponding concentration field is easier to solve for. We also consider a compactly supported tower source function given by
\begin{align} \label{eq:tower}
s(\bbx) = 
\left\{
\begin{array}{ll}
A  &  \text{if } \bbx \in \Omega_s    \\
0  &  \text{otherwise}    
\end{array}
\right. ,
\end{align}
where $\Omega_s \subset \Omega$ delineates the support and is set to $\Omega_s = [-0.4, -0.3] \times [-0.05, 0.05]$ for the following results. For consistency, we define a characteristic length $\gamma$ for this source as the side-length of its support. We adjust the source intensity $A$ for both sources to ensure that the resultant maximum concentration levels are close to unit. Note that for these arbitrary sources, we cannot solve the AD-PDE \eqref{eq:ADPDE} analytically. Instead, we use the FE solution as the ground-truth to calculate the error values according to \eqref{eq:err}.

\begin{figure} [t!]
        \centering
        \begin{subfigure}[b]{0.4\textwidth}
                \includegraphics[width=\textwidth]{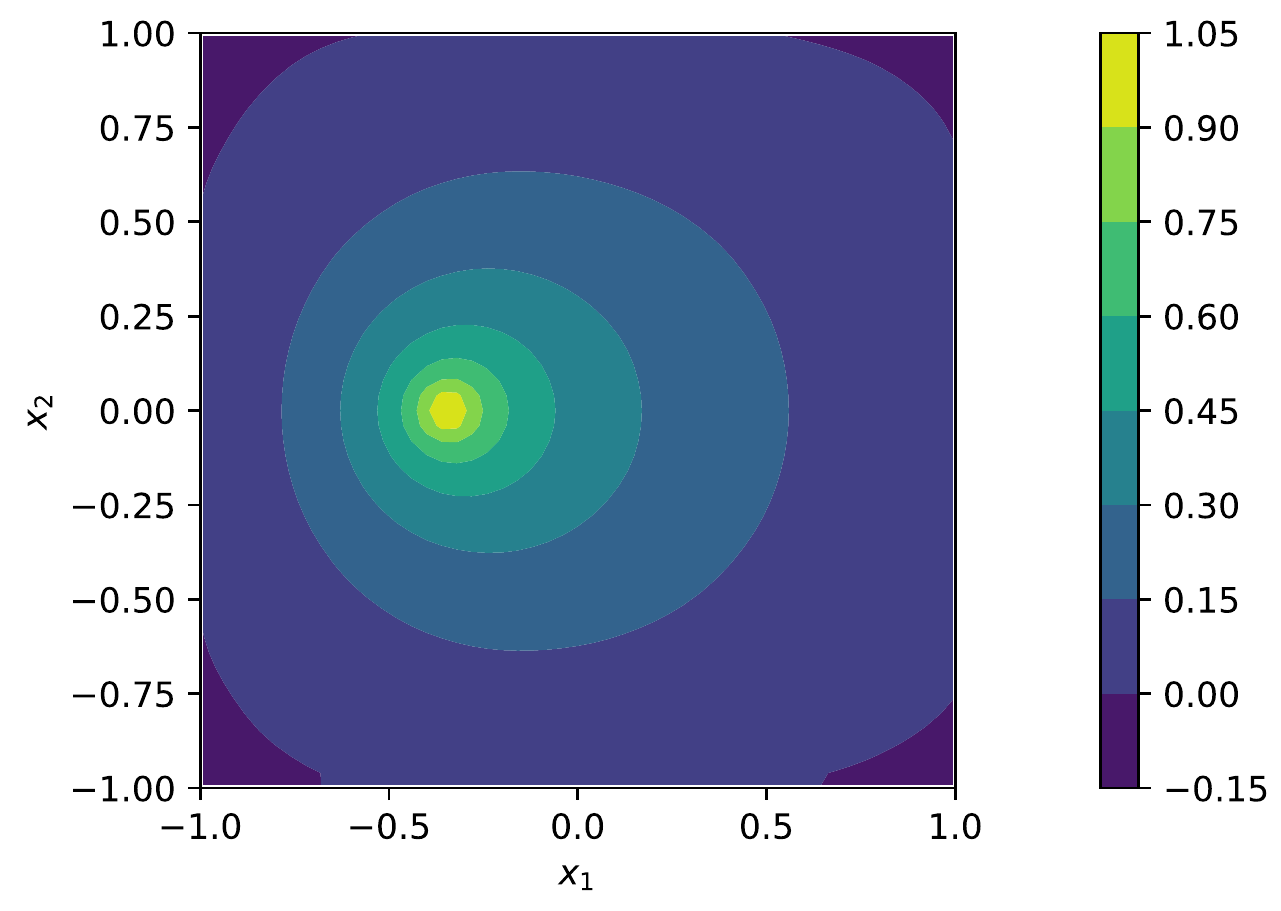}
    \caption{NN solution}	\label{fig:2D-cApp}
    \end{subfigure}
    \begin{subfigure}[b]{0.4\textwidth}
                \includegraphics[width=\textwidth]{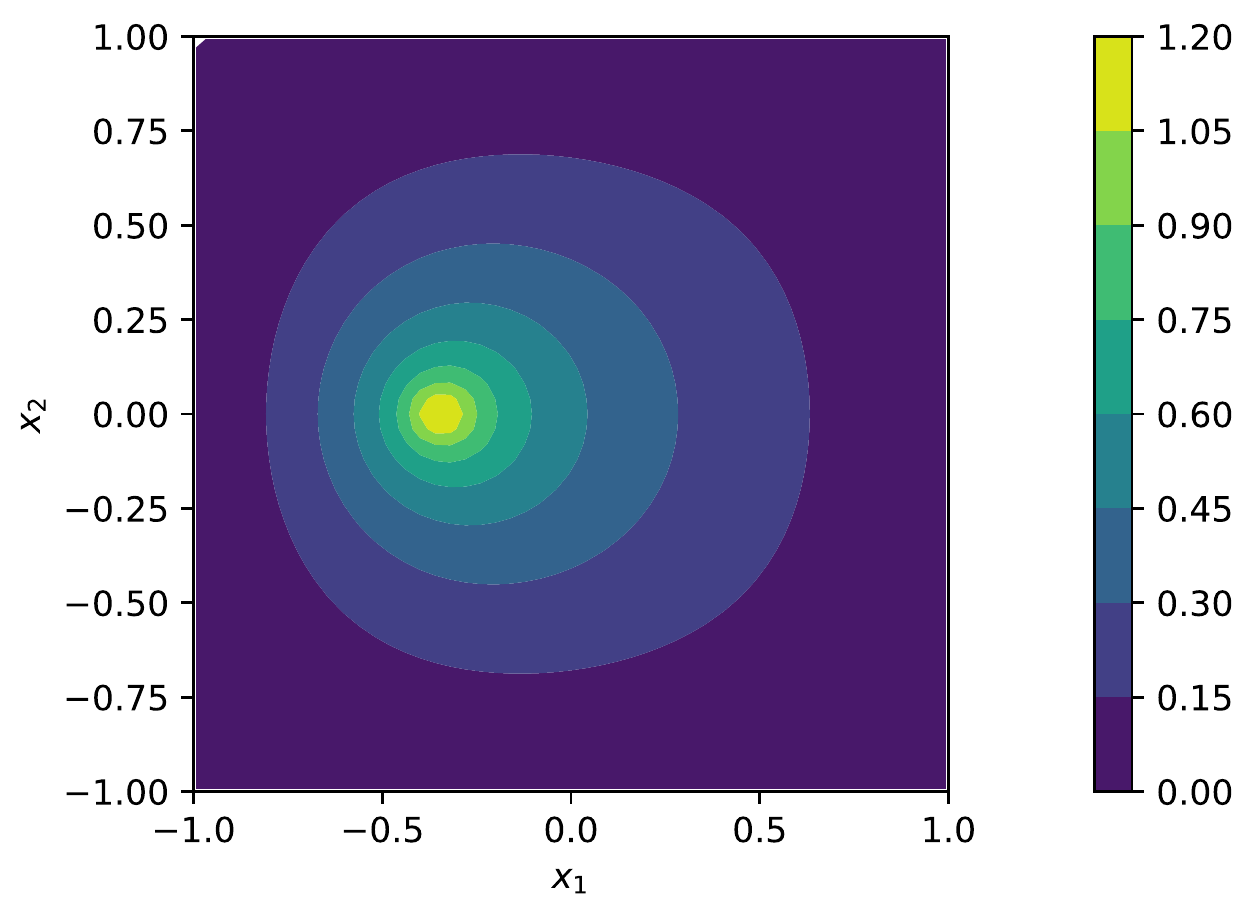}
    \caption{FE solution}	\label{fig:2D-cEx}
    \end{subfigure}
    \caption{NN and FE approximations of the concentration field for compactly-supported tower source corresponding to case 4 in Table \ref{table:err-2D}.}	\label{fig:2D-tower}
\end{figure}

In Table \ref{table:err-2D}, we report the performance of the \varnet Algorithm \ref{alg:VarNet} in solving the AD-PDE \eqref{eq:ADPDE} for these localized sources.
\begin{table}[t!]
\centering
\renewcommand{\arraystretch}{1.25}
\caption{Error values \eqref{eq:err} for the 2D AD problem with compactly supported source fields.}
\begin{tabular}{|c|c||c|c|c||c|} 
 \hline
No.	&	 $n$ 		& 	source	& 	$\gamma$	&	$n_v$		&	err 			\\ [0.5ex]  \hline\hline
1	&	 $81$	& Gaussian	& 	$0.2$		&	$50 \times 50$	&	0.00 			\\ 	 \hline
2	&	 $81$	& Gaussian	& 	$0.1$		&	$50 \times 50$	&	0.02 			\\ 	 \hline
3	&	 $271$	& Gaussian	& 	$0.05$		& $100 \times 100$	&	0.02 			\\ 	 \hline \hline
4	&	 $911$	& 	tower	& 	$0.1$		& $600 \times 600$	&	0.17 			\\ 	 \hline 
\end{tabular}
\label{table:err-2D}
\end{table}
Comparing the first three cases, observe that for smooth Gaussian sources, our proposed algorithm is capable of obtaining solutions identical to the FE method with a single layer MLP with 20 neurons. We set the weights to $\bbw = [10^4, 1]$ for this case. Nevertheless, the problem is more challenging when compactly supported tower source is used. Using a larger network with three layers $[10, 20, 30]$ amounting to $n=911$ trainable parameters, $n_v = 600 \times 600$ training points, and setting the weights to $\bbw = [10^4, 1]$, the algorithm solves the problem with err $= 0.17$.
Figure \ref{fig:2D-tower} shows the NN and FE approximations of the concentration field corresponding to case 4.
Observe that the general shape of the concentration field is recovered almost exactly but the peak value is not. Obviously, using larger network capacities $n$ and number of training points $n_v$ would decrease the error.

\subsection{2D AD-PDE with Turbulent Input Data} \label{sec:cvx}
\begin{figure}[t]
        \centering
	\begin{subfigure}[b]{0.38\textwidth}
		\includegraphics[width=\textwidth]{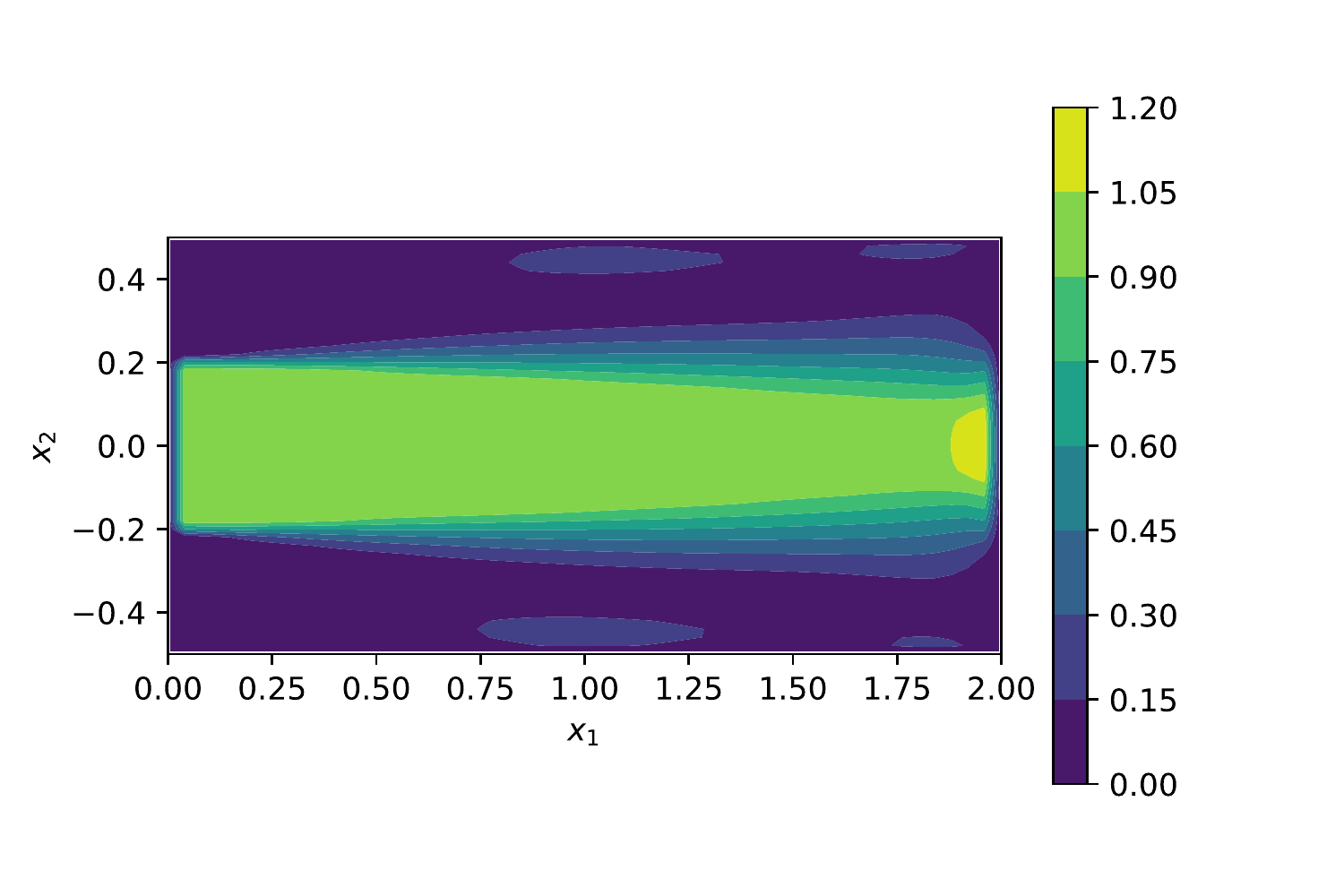}
	\caption{velocity magnitude field (m/s)} 	\label{fig:vel_cvx}
	\end{subfigure}
	\begin{subfigure}[b]{0.38\textwidth}
		\includegraphics[width=\textwidth]{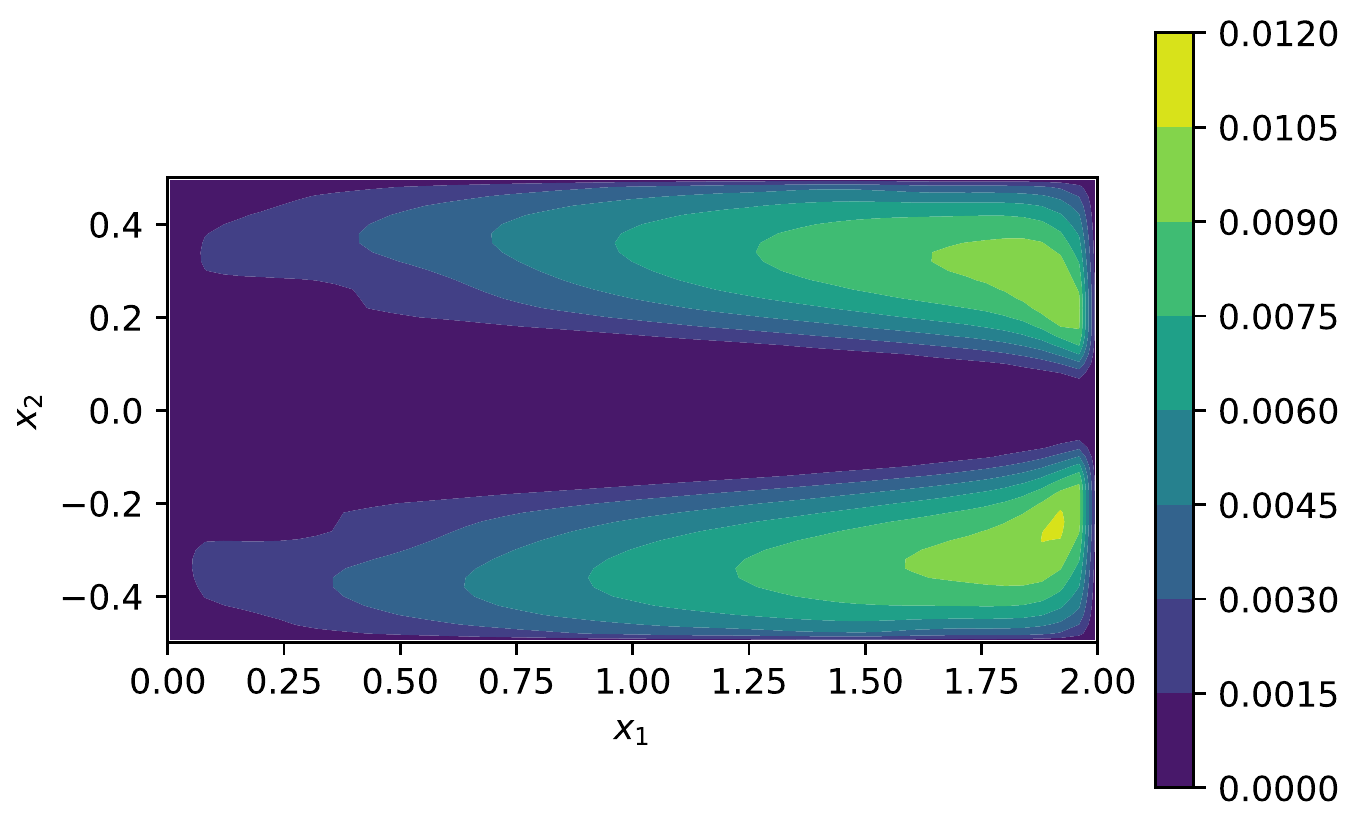}
	\caption{diffusivity field (m$^2$/s)}			\label{fig:diff_cvx}
	\end{subfigure}
	\caption{PDE input-data fields corresponding to the turbulent flow in Section \ref{sec:cvx} obtained using CFD techniques and \textsc{ANSYS-Fluent}.}	\label{fig:inpData_cvx}
\end{figure}

So far in the simulations, we have focused on simple flow fields that do not often occur in realistic problems. For instance, when AD-PDE \eqref{eq:ADPDE} is used to model mass transport in `air', even in very small velocities the resulting flow field is turbulent \cite{meJ3}, causing an excessive amount of mixing due to instantaneous changes in velocity vector. This often is captured with an equivalent turbulent diffusivity that is added to the laminar diffusivity of the medium; see \cite{meJ1} for more details.
In the following, we consider velocity and diffusivity fields that are obtained using computational fluid dynamics (CFD) methods via \textsc{ANSYS-Fluent}. Specifically consider the problem that we studied in Section \ref{sec:2Dt_anal} with the domain depicted in Figure \ref{fig:2Dt-domain} but in steady-state. We assume that air flows into this domain through boundary 2 with a velocity of 1m/s and leaves the domain through the opposite side. The corresponding velocity magnitude and diffusivity fields are given in Figure \ref{fig:inpData_cvx}.
Similar to Section \ref{sec:2Dt_anal}, we assume a constant Dirichlet BC on boundary 2, i.e., we set $g_2(\bbx) = 1$.\footnote{The concentration unit is arbitrary.} The rest of the BCs are set to zero.

\begin{figure}[t]
        \centering
	\begin{subfigure}[b]{0.38\textwidth}
		\includegraphics[width=\textwidth]{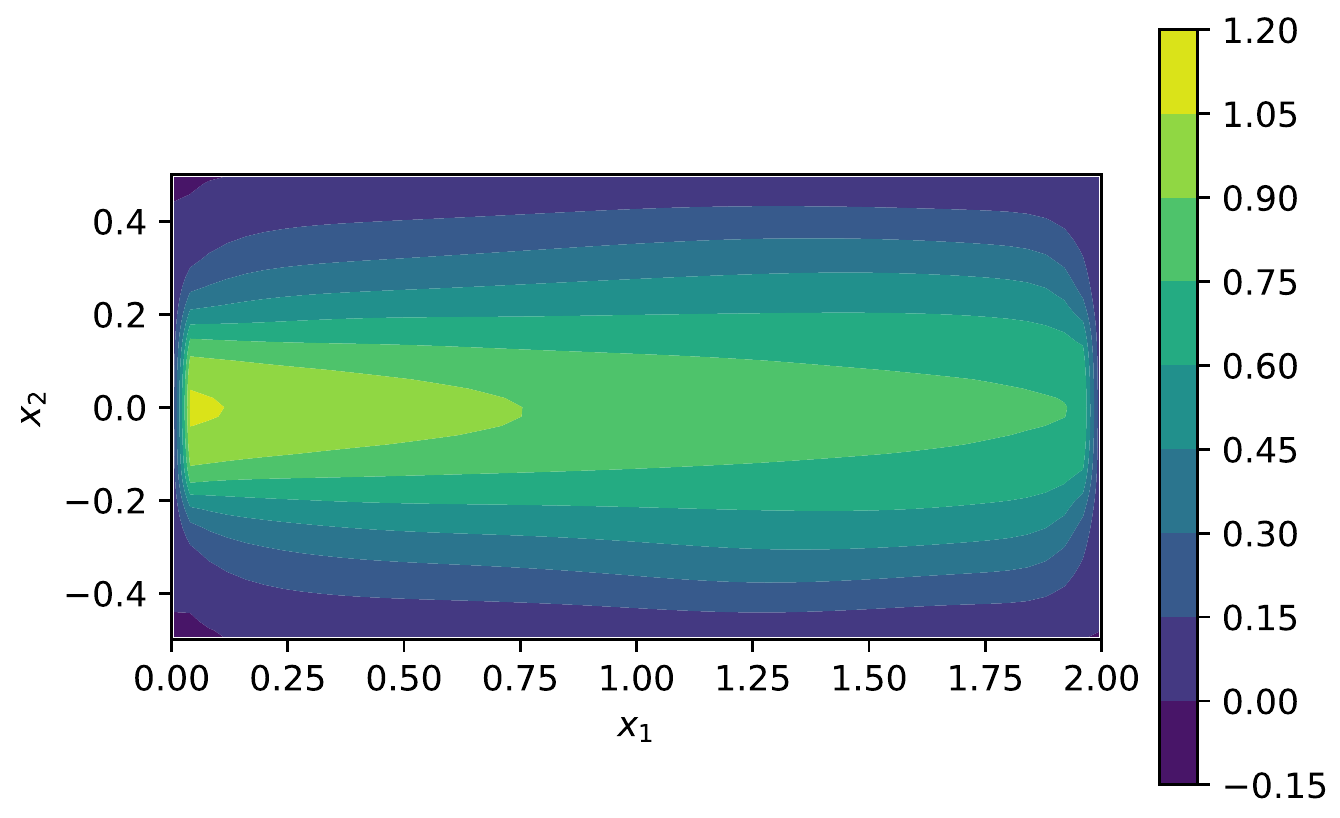}
	\caption{NN solution}			\label{fig:sim1-cApp}
	\end{subfigure}
	\begin{subfigure}[b]{0.38\textwidth}
		\includegraphics[width=\textwidth]{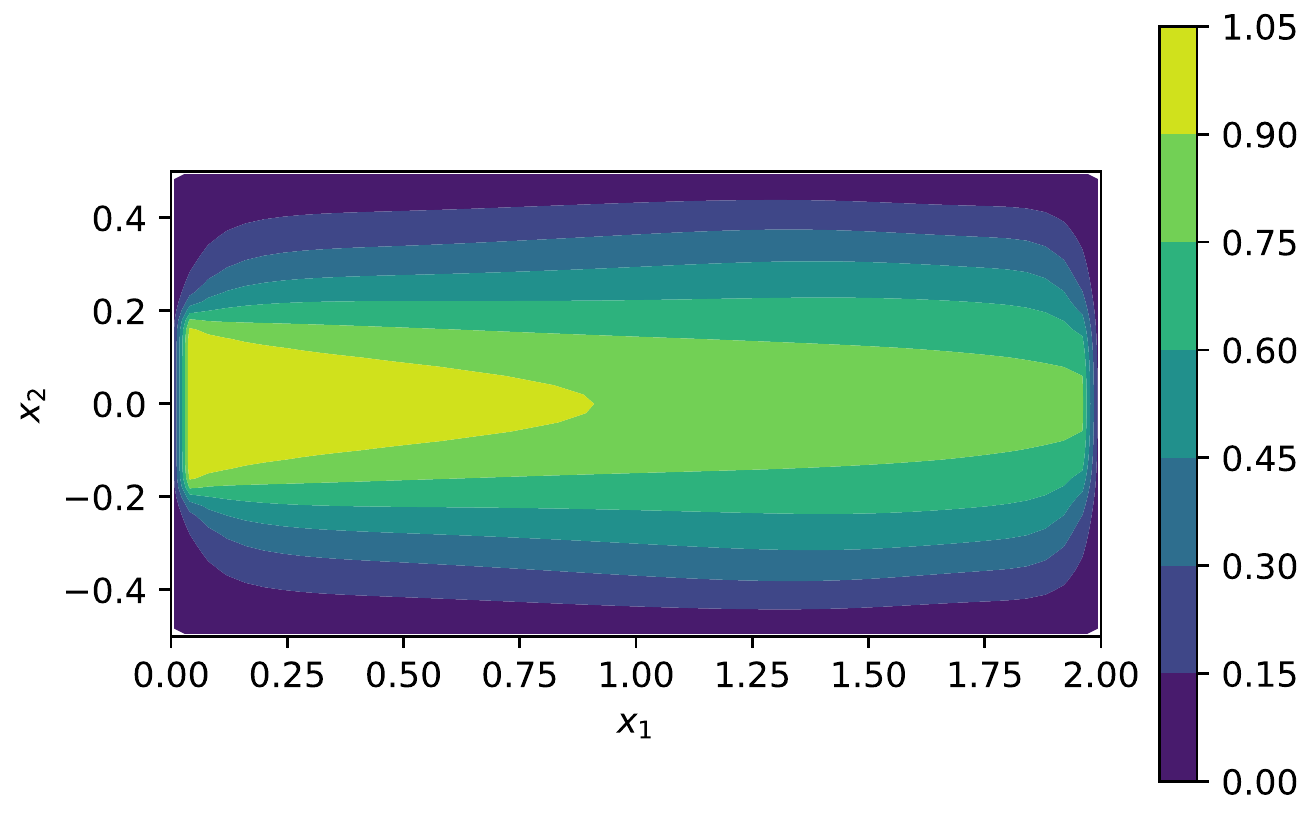}
	\caption{FE solution} 	\label{fig:sim1-cEx}
	\end{subfigure}
	\caption{NN and FE solutions for the AD problem of Section \ref{sec:cvx} with constant diffusivity field and turbulent velocity field depicted in Figure \ref{fig:vel_cvx}.}	\label{fig:sim1-sol}
\end{figure}

\begin{figure}[t]
	\centering
	\includegraphics[width=0.4\textwidth]{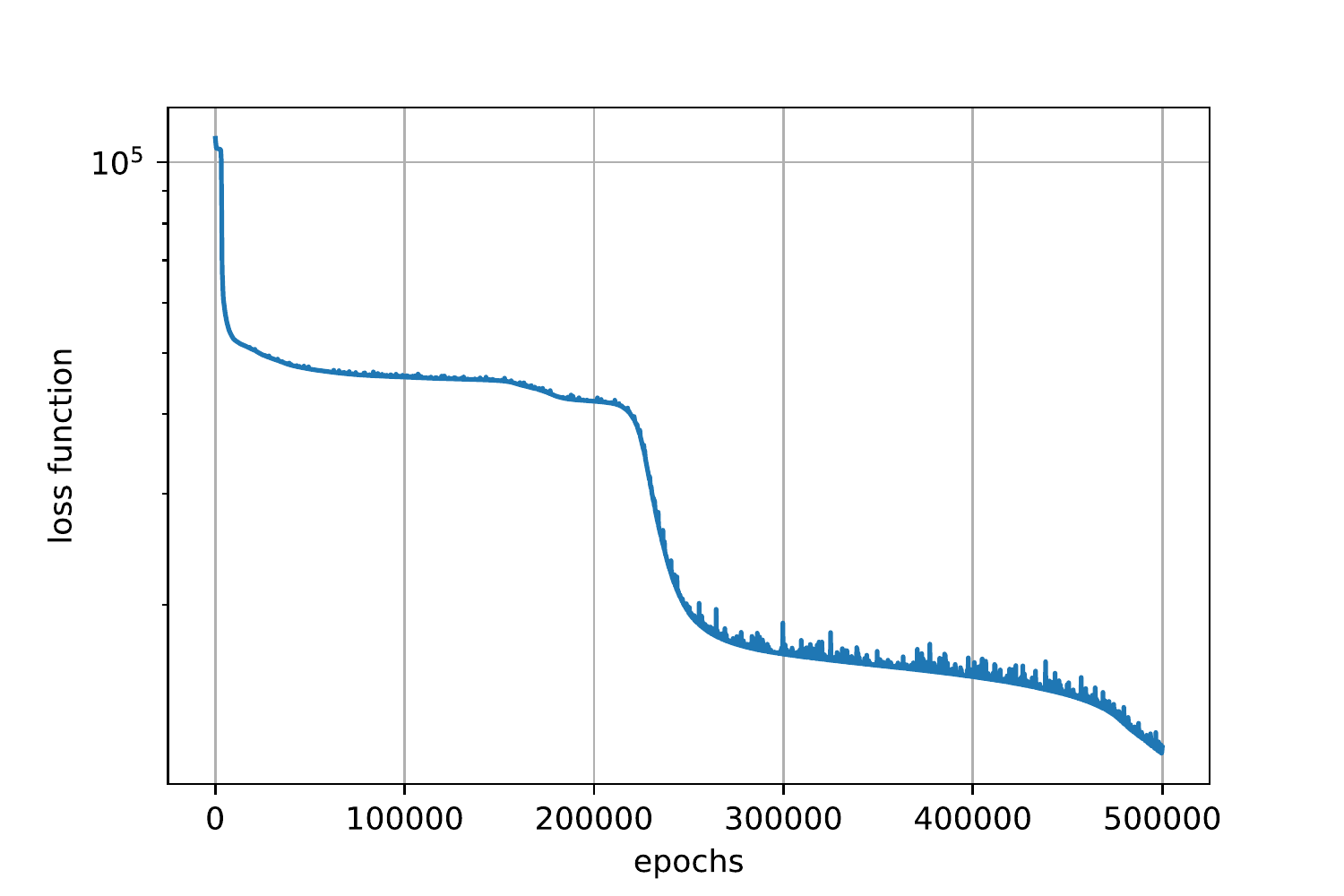}
	\caption{Loss function \eqref{eq:loss} as a function of training epochs for the AD problem of Section \ref{sec:cvx} with constant diffusivity and turbulent velocity fields.}	\label{fig:sim1-loss}
\end{figure}

First, we study the case of fixed diffusivity field by setting $\kappa = 0.01$ and using the non-constant velocity field of Figure \ref{fig:vel_cvx}. For this case, we use a MLP with three layers and $[10, 20, 30]$ neurons in the layers which amounts to $n=911$ trainable parameters. We set $\bbw = [1,1]$ and use $n_v = 400 \times 200$ training points. The approximation error for this case is err $=0.07$. Figure \ref{fig:sim1-sol} shows the approximate solutions obtained from \varnet Algorithm \ref{alg:VarNet} and the FE method, respectively.
Figure \ref{fig:sim1-loss} shows the loss value as a function of training epochs.
Note that for a large number of epochs the iterations get trapped in a local minimum after an initial sharp decrease and then start to drop again. In fact when the maximum number of epochs is reached, the solution is still improving. As more realistic, non-constant fields are used, the training process becomes more challenging since the loss function is more nonlinear. In such cases, larger number of epochs might be necessary to ensure convergence to an acceptable local minimum. Regular shuffling of the training data can also help with escaping such local minima.

Next, we consider the non-constant diffusivity field given in Figure \ref{fig:diff_cvx}. This results in an even more nonlinear loss function. To solve the AD-PDE \eqref{eq:ADPDE} in this case, we use a MLP with three layers and $[10, 20, 30]$ neurons in the layers which amounts to $n=911$ trainable parameters. We set $\bbw = [1,1]$ and use $n_v = 600 \times 300$ training points. The resulting approximation error for this case is err $=0.33$. Note that similar to Figure \ref{fig:2D-tower} in Section \ref{sec:compact}, the shape of the concentration field is reconstructed here but the peak value is inaccurate. Throughout the simulations, we have used very small MLP networks with small capacities. Using larger networks along with larger number of training points and training for longer periods will further improve the solutions. This is particularly true for this case since the loss curves obviously show that the NN is still improving when the maximum number of epochs is reached.

%% file: concl.tex
In this paper we proposed a new model-based unsupervised learning method, called \varnet, for the solution of partial differential equations (PDEs) using deep neural networks (NNs). Particularly, we proposed a novel loss function that relies on the variational (integral) form of PDEs as apposed to their differential form which is commonly used in the literature. Our loss function is discretization-free, highly parallelizable, and more effective in capturing the solution of PDEs since it employs lower-order derivatives and trains over measure non-zero regions of space-time. Given this loss function, we also proposed an approach to optimally select the space-time samples, used to train the NN, that is based on the feedback provided from the PDE residual. The models obtained using our algorithm are smooth and do not require interpolation. They are also easily differentiable and can directly be used for control and optimization of PDEs. Finally, the \varnet algorithm can straight-forwardly incorporate parametric PDE models making it a natural tool for model order reduction of PDEs. We demonstrated the performance of our method through extensive numerical experiments for the advection-diffusion PDE as an important case-study.

%% file: VarNet.tex
\subsection{Test Functions} \label{sec:testFun}
\begin{figure}[t]
    \centering
    \includegraphics[width=0.4\textwidth]{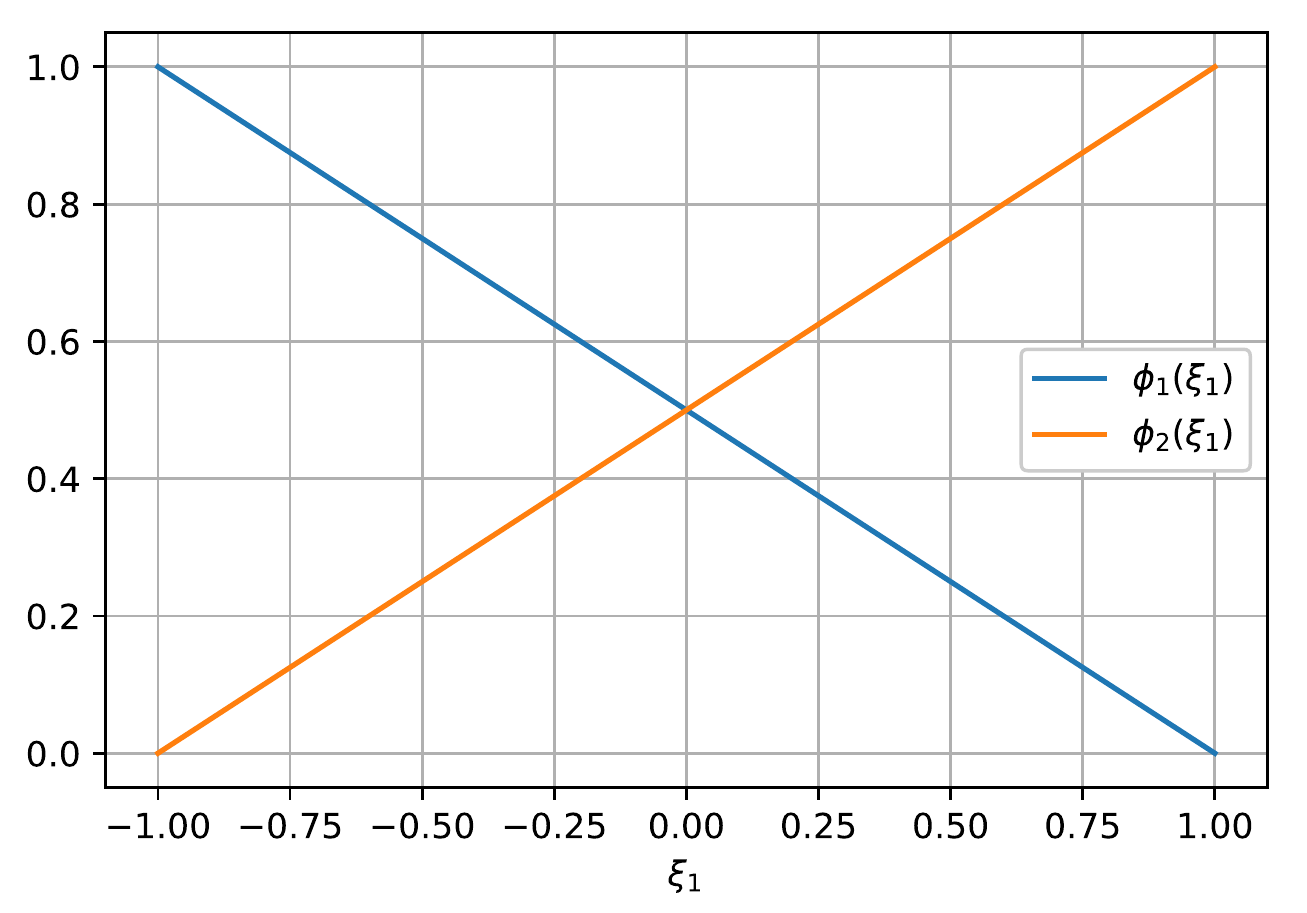}
    \caption{Linear shape functions for  the 1D element.}
    \label{fig:shapeF}
\end{figure}

As discussed in Section \ref{sec:loss}, the only requirement for the test function $v(t, \bbx)$ is that it must be compactly supported. Most of the FE shape functions satisfy this requirement and can be used with our proposed algorithm; see \cite{FEM2012H}.  Often in FE schemes standard elements are defined in the so called mathematical domain and then we can map these elements to the actual elements in the physical domain under isotropy assumption. In this paper we use a simple 1D standard element with linear shape functions as the building block to construct the desired test functions.
Particularly, let $\xi_1$ denote a generalized coordinate in the mathematical domain, that can be mapped to a spatial or temporal coordinate in the physical domain later on, and define the linear basis functions $\phi_1, \phi_2: [-1, 1] \to [0,1]$ given by
\begin{equation} \label{eq:shape1D}
\phi_1(\xi_1) = 0.5 \, (1-\xi_1) \and \phi_2(\xi_1) = 0.5 \, (1+\xi_1) .
\end{equation}
Figure \ref{fig:shapeF} shows the variation of these two shape functions along the length of the 1D element.

Given this 1D element, corresponding standard, higher-dimensional elements can easily be constructed. Let $\bbard$ denote the desired dimension of these elements where $\bbard = d+1$ for time-dependent problems and $\bbard=d$ otherwise. Furthermore, let $\xi_j \for 1\leq j \leq \bbard$ denote the coordinate along the $j$-th dimension, $\bbxi = (\xi_1, \dots, \xi_{\bbard}) \in \reals^{\bbard}$ denote the corresponding vector of coordinates, and the multi-index $(i_1, \dots, i_{\bbard})$ specify one of the $2^{\bbard}$ corners of the higher-dimensional element where $i_j \in \set{1,2}$.
Given this multi-index, the shape function corresponding to the respective corner of the element is defined by the multiplication of the 1D shape functions \eqref{eq:shape1D} as
\begin{equation} \label{eq:shapeF}
\bbarphi_{i_1, \dots, i_{\bbard}} (\bbxi) = \phi_{i_1}(\xi_1) \dots \phi_{i_{\bbard}}(\xi_{\bbard}) .
\end{equation}

For simplicity of presentation, let the augmented coordinate $\barbx \in \reals^{\bbard}$ be defined as $\barbx = (t, \bbx)$ for time-dependent problems and $\barbx = \bbx$, otherwise. Then, we denote the shape functions in the physical domain by $N_{i_1, \dots, i_{\bbard}}(\barbx)$.
Assuming isotropic elements, we can write the coordinates in the physical domain in terms of the coordinates in the mathematical domain as
\begin{equation} \label{eq:isotropic}
\bbarx_j = \sum\nolimits_{i_1, \dots, i_{\bbard}} \bbarx_{j; \, i_1, \dots, i_{\bbard}} \ \bbarphi_{i_1, \dots, i_{\bbard}} (\bbxi) .
\end{equation}
In words, given the coordinates $\bbxi$ in the mathematical domain, the corresponding $j$-th coordinate $\bbarx_j$ in the physical domain is given as a convex combination of the $j$-th coordinates $\bbarx_{j; i_1, \dots, i_{\bbard}}$ of the corners of the element, specified by the multi-index $(i_1, \dots, i_{\bbard})$.
This mapping from mathematical to physical domain, implicitly defines the shape functions $N_{i_1, \dots, i_{\bbard}}(\barbx)$ as
$$ N_{i_1, \dots, i_{\bbard}}(\barbx) = \bbarphi_{i_1, \dots, i_{\bbard}}(\bbxi) . $$
Without loss of generality, we use cubical elements so that this mapping amounts to simple scalings $h_j >0$ of the coordinates where $1\leq j \leq \bbard$. Note that this simple choice of elements considerably facilitates the implementation of the VarNet Algorithm \ref{alg:VarNet}. Nevertheless, the type, shape, and size of the elements are arbitrary and can be as general as the ones used in different FE schemes \cite{FEM2012H}.

\begin{figure}[t]
    \centering
    \includegraphics[width=0.4\textwidth]{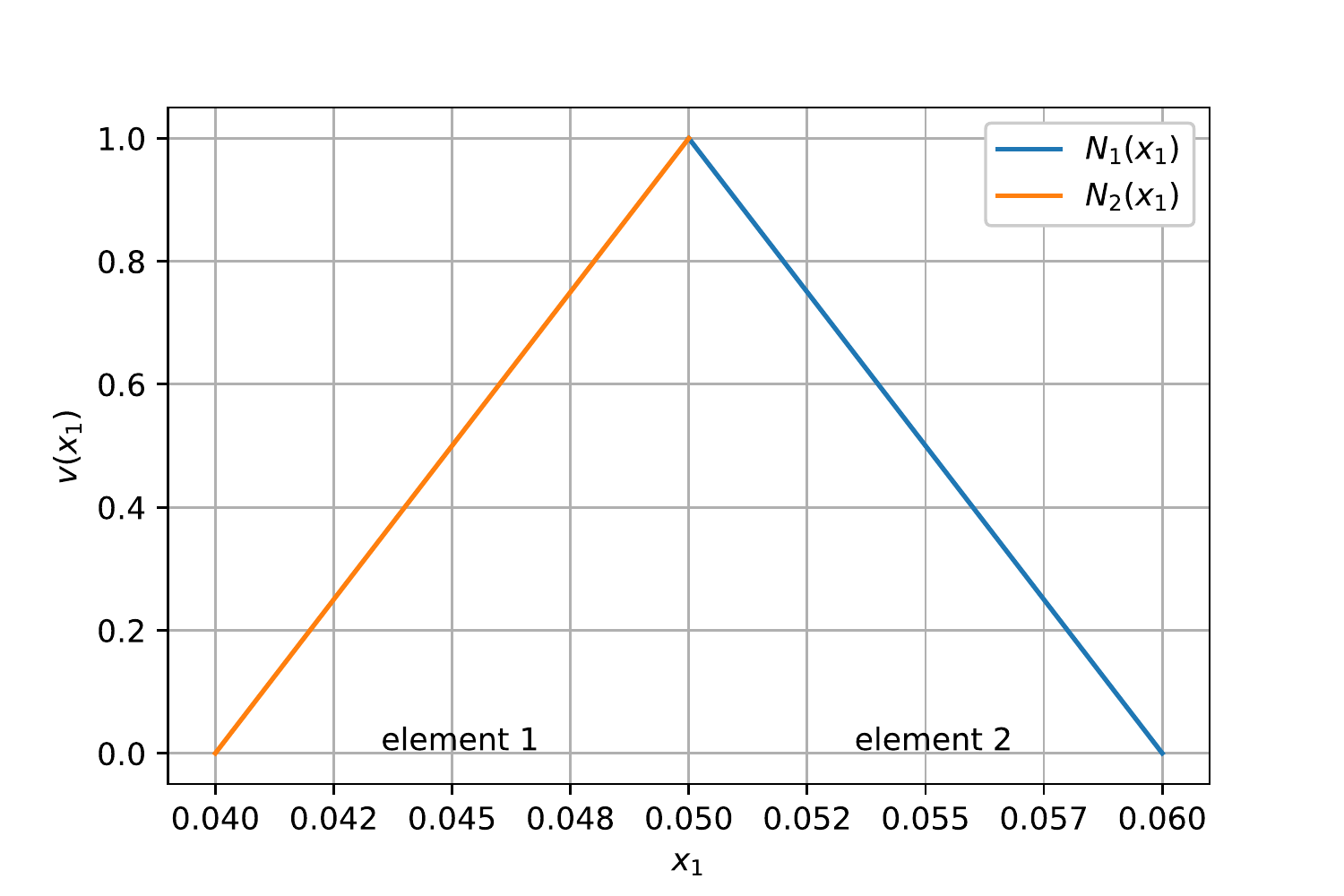}
    \caption{A test function for the 1D physical domain. The support of the function is compact and contains two 1D elements with length $0.01$ that share the training point located at $x_1 = 0.05$.}
    \label{fig:testF}
\end{figure}

Given the above shape functions, we are ready to define the test functions that we use in the loss function \eqref{eq:loss}. Specifically, given a training point in the space-time, we consider a test function $v(\barbx)$ centered at that point that belongs to a set of elements that share the point. Then, we define the desired test function as $v(\barbx) = N_{i_1, \dots, i_{\bbard}}(\barbx)$ over each of these elements where the multi-index $(i_1, \dots, i_{\bbard})$ corresponds to the shared corner of that element.
As an example, the test function for the 1D (physical) domain corresponding to Figure \ref{fig:shapeF}, centered at the training point $x_1=0.05$ with element scaling of $h_1 = 0.01$, is depicted in Figure \ref{fig:testF}.
Over element 1, the test function is defined as $v(x_1) = N_2(x_1)$ whereas over element 2, it is defined as $v(x_1) = N_1(x_1)$. Note that the support of $v(x_1)$ is obtained by patching together elements 1 and 2 and is compact.
During training, we arbitrarily select the location of the test functions over the space-time while fixing the shape and size of the elements. As mentioned earlier, this assumption can be relaxed at the expense of more computational cost.

\subsection{Numerical Computation of Loss Function} \label{sec:lossComp}
In order to compute the loss function \eqref{eq:loss}, we need to compute the variational form \eqref{eq:varPDE} which in turn requires the calculation of the derivatives of the test function $v(\barbx)$. To this end, we first derive the derivatives of the shape function $N_{i_1, \dots, i_{\bbard}}(\barbx)$ for one of the elements within the support of $v(\barbx)$. Then, the desired derivatives of $v(\barbx)$ are obtained from the derivatives of these shape functions over each element. The decomposed form of the shape function \eqref{eq:shapeF} makes its differentiation straight-forward. Specifically, using \eqref{eq:shape1D} for coordinate $\xi_j$ in the mathematical domain we have
\begin{equation} \label{eq:shapeDeriv}
\frac{\partial \bbarphi_{i_1, \dots, i_{\bbard}} }{ \partial \xi_j } = \mp 0.5 \, \phi_{i_1}(\xi_1) \dots \phi_{i_{j-1} }(\xi_{j-1}) \phi_{i_{j+1} }(\xi_{j+1}) \dots \phi_{i_{\bbard}}(\xi_{\bbard}) ,
\end{equation}
where minus sign corresponds to $i_j=1$.
Let the matrix $\nabla \Phi \in \reals^{\bbard \times 2^{\bbard}}$ collect these derivatives where each column corresponds to one shape function, specified by the multi-index $(i_1, \dots, i_{\bbard})$, and is equal to the gradient $\nabla \bbarphi_{i_1, \dots, i_{\bbard}}$ whose entries are given by \eqref{eq:shapeDeriv}.
Furthermore, define the Jacobian matrix $\bbJ \in \reals^{\bbard \times \bbard}$ whose column $j$ is the gradient $\nabla \bbarx_j$ of the coordinate $\bbarx_j$ with respect to the mathematical coordinates.
Let also the matrix $\bbX \in \reals^{2^{\bbard} \times \bbard }$ collect in its column $j$, the $j$-th coordinate of the corners of the element in the order specified by the multi-index $(i_1, \dots, i_{\bbard})$. Then, we have
$$ \bbJ = \nabla \Phi \cdot \bbX . $$
This expression can be checked by direct differentiation of \eqref{eq:isotropic}. For the cubical elements that we use in the physical domain, this Jacobian matrix is given explicitly by
\begin{equation} \label{eq:Jacob}
\bbJ = \frac{1}{2} \, \diag(h_1, \dots, h_{\bbard}) ,
\end{equation}
where $h_j$ denotes the scaling along coordinate $j$ from mathematical to physical domain.
Finally, let the matrix $\bbB \in \reals^{\bbard \times 2^{\bbard}}$ collect the desired derivatives $\nabla_{\barbx} N_{i_1, \dots, i_{\bbard}}$, with respect to the coordinates $\barbx$, in its columns. Then, we have
$$ \bbB = \bbJ^{-1} \nabla \Phi . $$
The \code{FE} class in the \varnet tool \cite{VarNet} computes the test functions, presented in the previous section, as well as their derivatives as discussed above.

The standard elements, defined in the mathematical domain, make the numerical computation of the variational form \eqref{eq:varPDE} and the loss function \eqref{eq:loss} very efficient. This is because the Gauss-Legendre quadrature rules can be used for integration over these elements.
Specifically, let $\bbarn$ denote the number of integration points for the 1D element introduced in Section \ref{sec:testFun}, $\bbarxi_i$ denote the values of the coordinate $\xi_1$ at these integration points for $i \in \set{1, \dots, \bbarn}$, and $\bbarw_i$ denote the corresponding weights for a 1D $\bbarn$-point quadrature rule. For instance for $\bbarn = 2$, we have $\bbarxi_i = \pm 1/\sqrt{3}$ and $\bbarw_i = 1, \forall i$.
Let $\bbarf: \reals^{\bbard} \to \reals$ denote the integrand in \eqref{eq:varPDE}.
Given the 1D integration point information, we can approximate the variational form of the PDE \eqref{eq:varPDE} by a $\bbard$-dimensional quadrature rule as
\begin{align*}
\bbarl &= \int_{\Omega} \bbarf(\barbx) \, d\barbx = \int_{\bbarOmega} \bbarf(\barbx(\bbxi)) \, \abs{\bbJ} d\bbxi = \abs{\bbJ} \int_{\bbarOmega} \bbarf(\barbx(\bbxi)) \, d\bbxi \\
& \approx \abs{\bbJ} \sum\nolimits_{i_1, \dots, i_{\bbard}} \bbarw_{i_1} \dots \bbarw_{i_{\bbard}} \, \bbarf(\barbx(\bbarxi_{i_1}, \dots, \bbarxi_{i_{\bbard} } )) ,
\end{align*}
where $\abs{\bbJ}$ denotes the determinant of the Jacobian matrix \eqref{eq:Jacob} which is constant for the cubical elements used here, $\bbarOmega$ denotes the compact support of the test function $v(\bbx)$, and $i_j \in \set{1, \dots, \bbarn}$ for $1 \leq j \leq \bbard$.
For small element sizes, the integrand $\bbarf(\bbx)$ can accurately be estimated by low order polynomials making the numerical integration very accurate. Finally, note that the integrand $\bbarf(\bbx)$ involves the PDE input data that need to be evaluated at the integration points. Since these integration points are defined in the mathematical coordinates, we map them to the physical domain using \eqref{eq:isotropic}.